\documentclass{article}
\usepackage{ijcai23}

\usepackage{times}
\usepackage{soul}
\usepackage{url}
\usepackage[hidelinks]{hyperref}
\usepackage[utf8]{inputenc}
\usepackage[small]{caption}
\usepackage{graphicx}
\usepackage{amsmath}
\usepackage{amsthm}
\usepackage{booktabs}
\usepackage{algorithm}
\usepackage{algorithmic}
\usepackage[switch]{lineno}
\usepackage{amssymb}
\usepackage{xcolor}
\usepackage{subcaption}
\usepackage{multirow}

\newcommand{\citet}[1]
{\citeauthor{#1}~\shortcite{#1}}

\usepackage{cleveref} %

\newtheorem{definition}{Definition}

\title{Multi-Robot Coordination and Layout Design for Automated Warehousing\footnote{The paper can be found in the IJCAI 2023 proceeding at \url{https://www.ijcai.org/proceedings/2023/0611}}}

\author{
Yulun Zhang$^1$
\and
Matthew C. Fontaine$^2$\and
Varun Bhatt$^2$\and
Stefanos Nikolaidis$^2$\And
Jiaoyang Li$^1$
\affiliations
$^1$Robotics Institute, Carnegie Mellon University\\
$^2$Department of Computer Science, University of Southern California
\emails
yulunzhang@cmu.edu,
\{mfontain,vsbhatt,nikolaid\}@usc.edu,
jiaoyangli@cmu.edu
}

\begin{document}

\maketitle

\begin{abstract}

With the rapid progress in Multi-Agent Path Finding (MAPF), researchers have studied how MAPF algorithms can be deployed to coordinate hundreds of robots in large automated warehouses. While most works try to improve the throughput of such warehouses by developing better MAPF algorithms, we focus on improving the throughput by optimizing the warehouse layout. We show that, even with state-of-the-art MAPF algorithms, commonly used human-designed layouts can lead to congestion for warehouses with large numbers of robots and thus have limited scalability. We extend existing automatic scenario generation methods to optimize warehouse layouts. Results show that our optimized warehouse layouts (1) reduce traffic congestion and thus improve throughput, (2) improve the scalability of the automated warehouses by doubling the number of robots in some cases, and (3) are capable of generating layouts with user-specified diversity measures. We include the source code at: \url{https://github.com/lunjohnzhang/warehouse_env_gen_public}

\end{abstract}

\section{Introduction}

Today, hundreds of robots are navigating autonomously in warehouses to transport goods from one location to another~\cite{WurmanAAAI07,KouAAAI20}. Such automated warehouses are currently a multi-billion-dollar industry led by big companies like Amazon, Alibaba, and Ocado. To improve the throughput of automated warehouses, a lot of works have studied the underlying Multi-Agent Path Finding (MAPF) problem for coordinating warehouse robots smoothly~\cite{NguyenIJCAI17,MaAAMAS17,LiuAAMAS19,LiAAMAS20a,KouAAAI20,ChenRAL21,ContiniRAS21,VaramballySoCS22,DamaniRAL21}.

\begin{figure}
    \centering

    \begin{subfigure}{0.23\textwidth}
      \centering
      \includegraphics[width=.97\textwidth]{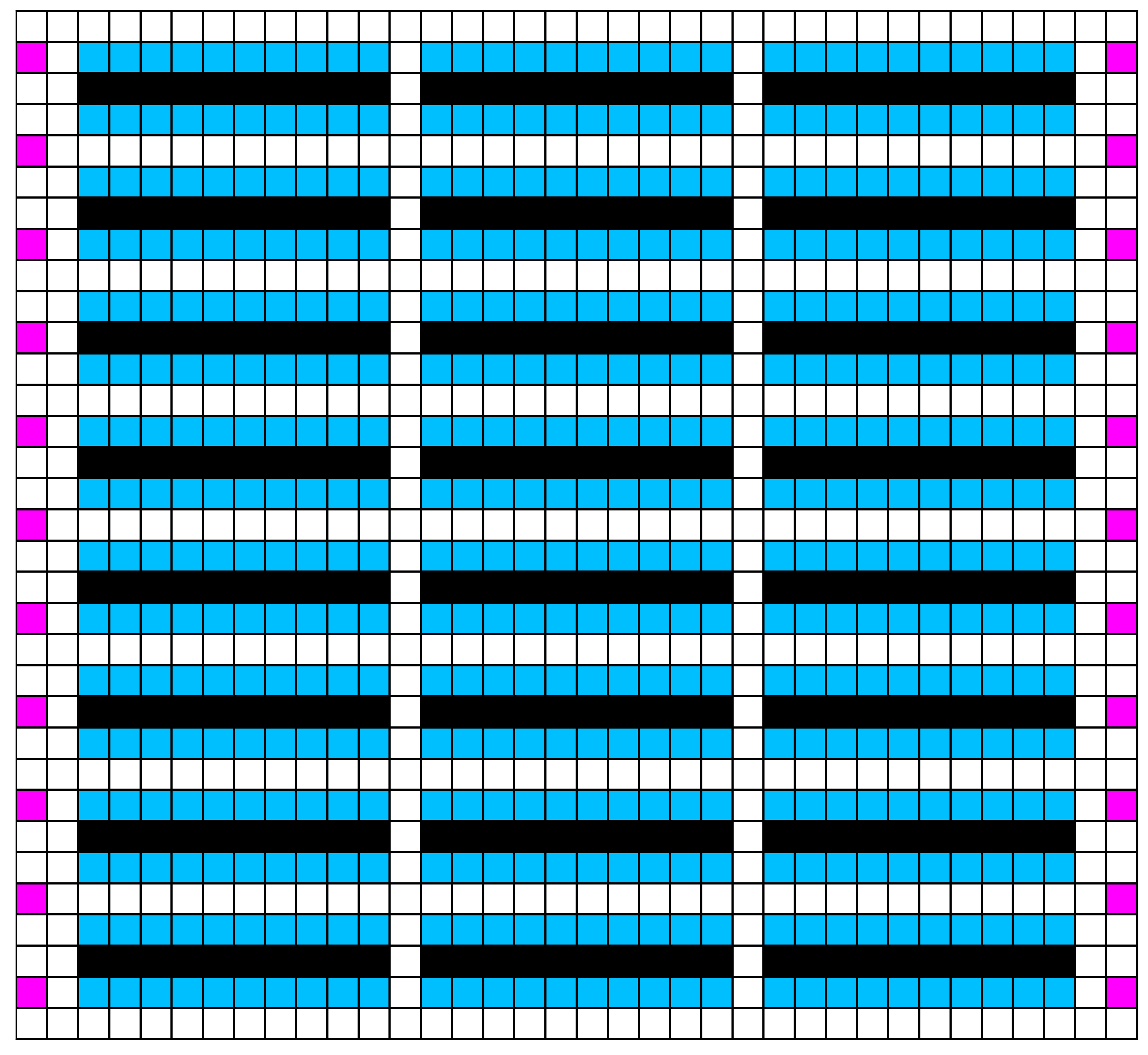}
      \caption{Human-designed}
      \label{fig:front-fig:human}
    \end{subfigure}%
    \hfill
    \begin{subfigure}{0.23\textwidth}
      \centering
      \includegraphics[width=.97\textwidth]{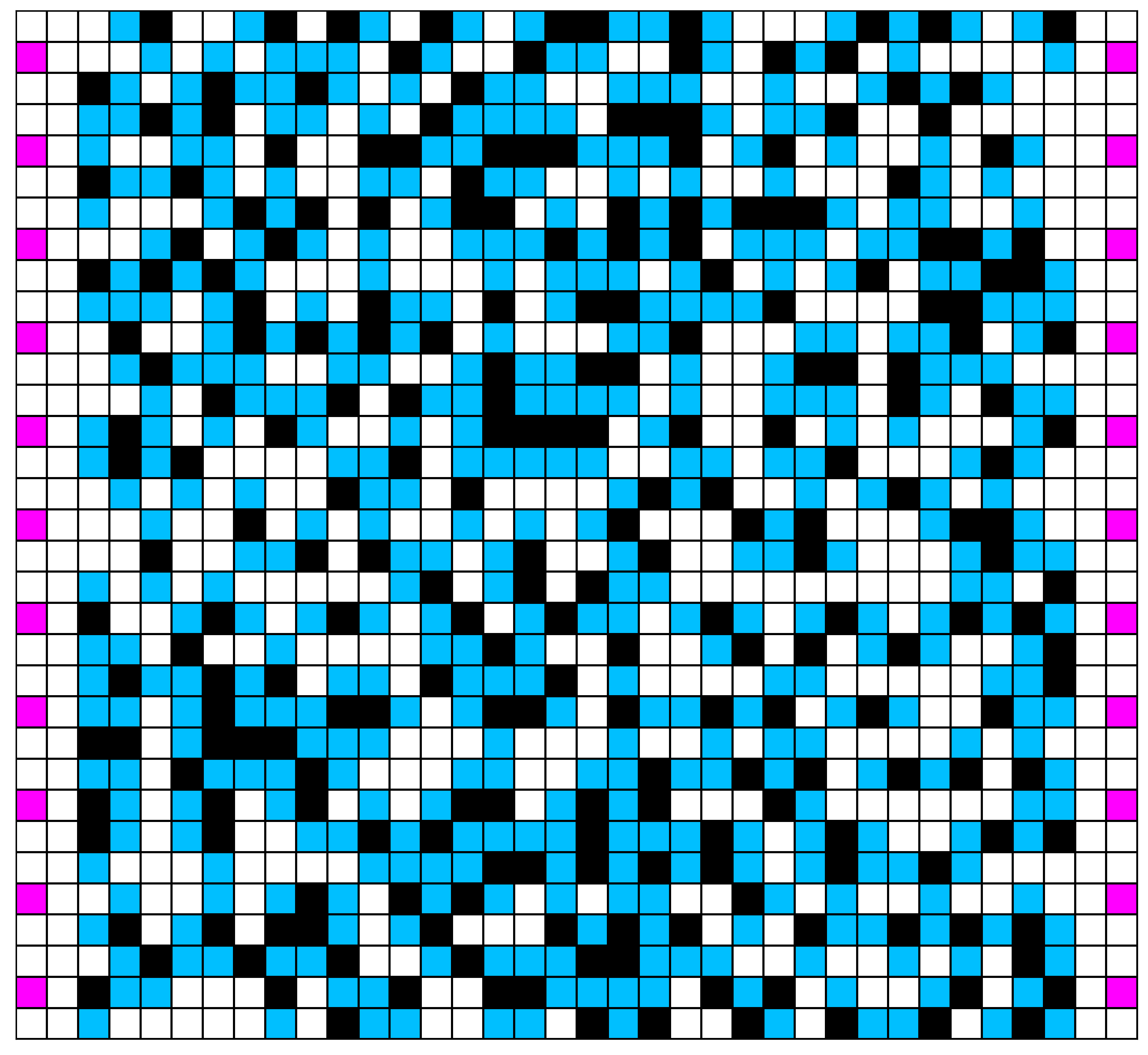}
      \caption{Optimized}
      \label{fig:front-fig:me-opt}
    \end{subfigure}%

    \caption{%
    Example of layout optimization. %
    (a) and (b) illustrate a commonly used human-designed layout and our optimized layout, respectively. Robots move to blue and pink tiles alternatively to transport goods. Black tiles are shelves, which the robots cannot traverse. %
    }
    \label{fig:front-fig}
\end{figure}

These works always use human-designed layouts, with an example shown in \Cref{fig:front-fig:human} and in \Cref{appen:human-layout}, to evaluate MAPF algorithms for automated warehouses.  These human-designed layouts originated from the layouts for traditional warehouses where human workers, instead of robots, transport goods. They usually follow regularized patterns with shelves clustered in rectangular shapes. Such patterns make it easy for human workers to locate goods. But, for robots, it hardly matters to have such regularized patterns. We thus abandon pattern regularity to explore novel layouts with better throughput. \Cref{fig:front-fig:me-opt} shows one of our optimized layouts with the same number of shelves.

Therefore, instead of developing better MAPF algorithms, we propose to improve the throughput of automated warehouses by optimizing warehouse layouts. %
We use the Quality Diversity algorithm 
MAP-Elites~\cite{mouret2015illuminating} to optimize the warehouse layouts with the objective of maximizing the throughput produced by a MAPF-based robot simulator while diversifying a set of user-defined measures, such as travel distances of the robots and the distribution of the shelves. In case the layout found by MAP-Elites is invalid, we follow previous work~\cite{zhang:aiide2020,fontaine2021importance} and use a Mixed Integer Linear Programming (MILP) solver to repair it. Since both the robot simulator and the MILP solver are time-consuming, we combine them with an online trained surrogate model %
using Deep Surrogate Assisted Generation of Environments (DSAGE)~\cite{Bhatt2022DeepSA}. %

We make the following contributions:  (1) we show that, even with state-of-the-art MAPF algorithms, commonly used human-designed layouts can lead to congestion for warehouses with large numbers of robots, (2) we propose the first layout optimization method for automated warehouses based on MAP-Elites and DSAGE, (3) we extend DSAGE by scaling it from single-agent domains to the warehouse domain with up to 200 robots and incorporating MILP to ensure that the layouts satisfy warehouse-specific constraints, (4) we show that our optimized layouts greatly reduce the traffic congestion and improve the throughput compared to commonly used human-designed layouts, and (5) we show that our proposed layout optimization method can generate a diverse set of high throughput layouts with respect to user-defined diversity measures.

\section{Problem Definition}

\begin{figure}[!t]
    \centering
    \begin{subfigure}[t]{0.2\textwidth}
      \centering
    \includegraphics[width=1\textwidth]{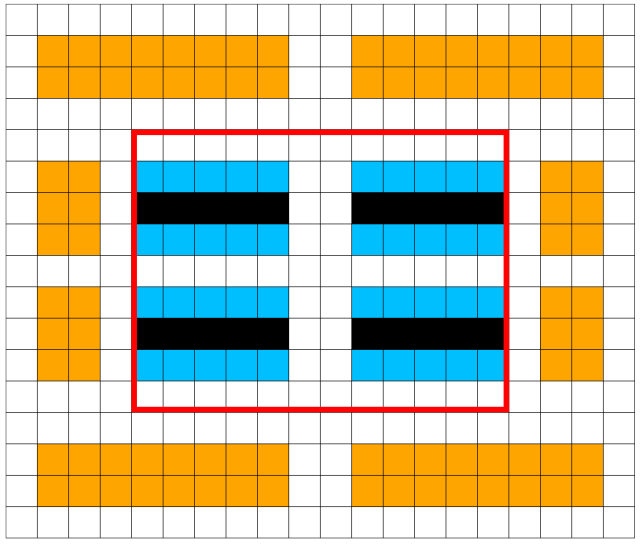}
      \caption{Home-location scenario}
      \label{fig:exp-setup-r}
    \end{subfigure}%
    \hfill
    \begin{subfigure}[t]{0.25\textwidth}
      \centering
    \includegraphics[width=0.8\textwidth]{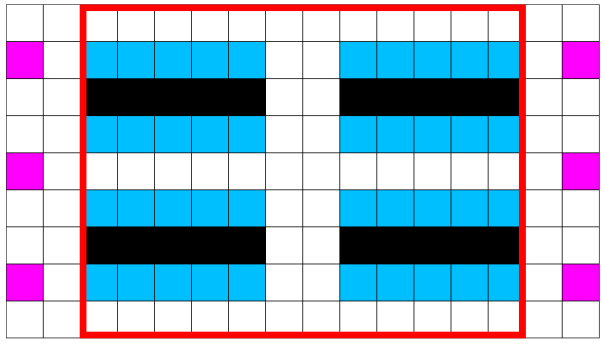}
      \caption{Workstation scenario}
      \label{fig:exp-setup-w}
    \end{subfigure}
    \caption{Example layouts. We optimize tiles inside the red box.}
    \label{fig:exp-setup}
\end{figure}

We formally define the warehouse layouts as follows with two examples shown in \Cref{fig:exp-setup}. Following the terminology in the MAPF community, we use agents to refer to robots.
\begin{definition}[Warehouse layout]
We represent the warehouse layout with a four-neighbor grid, where each tile can be one of the following five types. Black tiles represent shelves to store packages. Blue tiles represent endpoints where agents park and interact with the adjacent shelves. Pink tiles represent workstations where agents interact with human workers. White tiles represent empty spaces. Orange tiles represent home locations for agents, which are empty spaces with special properties used by some MAPF algorithms. Agents can traverse non-black tiles.
\end{definition}

Since the orange, blue, and pink tiles are the potential start and goal locations of the agents, a valid warehouse layout needs to make these tiles connected to each other. Furthermore, since the agents interact with the shelves from the endpoints, a valid warehouse layout needs to also make blue and black tiles next to each other.
\begin{definition}[Valid layout]
A warehouse layout is valid iff (1) any two tiles of blue, pink, or orange color are connected through a path with non-black tiles, (2) each blue tile is adjacent to at least one black tile, and (3) each black tile is adjacent to at least two blue tiles.
\end{definition}

We further define the well-formed property~\cite{LiuAAMAS19} for warehouse layouts, a required property by some MAPF algorithms.

\begin{definition}[Well-formed layout]
A valid warehouse layout is well-formed iff the number of orange tiles is no smaller than the number of agents, and any two tiles of blue or orange color are connected through a path with only white tiles. 
\end{definition}

We pick two common warehouse scenarios to test our layout optimization methods, one requiring well-formed layouts and one requiring valid layouts.

\paragraph{Home-Location Scenario}
Following previous work~\cite{LiuAAMAS19,Li2020LifelongMP}, the home-location scenario simulates a warehouse in which the agents transport goods from one shelf to another. That is, the agents are asked to move from one blue tile to another.
\Cref{fig:exp-setup-r} shows an example. %
We require the layouts %
to be well-formed. 

\paragraph{Workstation Scenario} 
Following previous work~\cite{Hnig2019PersistentAR}, the workstation scenario simulates a warehouse in which the agents transport goods between workstations and shelves. That is, the agents are asked to move to pink and blue tiles alternately. 
\Cref{fig:exp-setup-w} shows an example. %
We require the layouts %
to be valid. %

Finally, we define our optimization objective \emph{throughput} and our layout optimization problem. Our goal is to find novel distributions of the shelves and their associated endpoints. We thus keep the \textit{non-storage area}, i.e., area outside of the red box in \Cref{fig:exp-setup}, unchanged for simplicity and only optimize the \textit{storage area}, i.e., area inside the red box.

\begin{definition}[Throughput]
    An agent finishes a task when it reaches its current assigned goal location. The throughput is the average number of finished tasks per timestep.
\end{definition}

\begin{definition}[Layout Optimization]\label{def:layout-opt}
    Given the layout of a non-storage area and a desired number of shelves $N_s$, the layout optimization problem searches for the allocation of black, blue, and white tiles inside the storage area to find valid/well-formed layouts with $N_s$ black tiles while maximizing their throughput and diversifying their user-defined diversity measures.
\end{definition}

\section{Preliminaries} \label{sec:pre}

We now introduce the (lifelong) MAPF algorithms that we use in our simulator and existing works on warehouse layout optimization and scenario generation with Quality Diversity algorithms.

\subsection{Lifelong Multi-Agent Path Finding}

Multi-Agent Path Finding (MAPF)~\cite{SternSoCS19} aims to find collision-free paths for a group of agents from their given start locations to goal locations. Each agent has only one pair of start and goal locations and, after it reaches its goal location, waits there forever. Lifelong MAPF~\cite{Li2020LifelongMP} is a variant of MAPF in which the agents do not stop and wait at their goal locations. Instead, a new goal location is assigned once an agent reaches its current goal location so that the agents keep moving constantly. 
In this work, we use a widely-used lifelong MAPF definition~\cite{Li2020LifelongMP} where (1) the agents move on a four-neighbor grid, (2) in each discrete timestep, every agent can either stay at its current tile or move to an adjacent tile, (3) the agents cannot be at the same tile or swap their locations at adjacent tiles at the same timestep, and (4) the goal is to maximize the throughput. %

Previous works have approached the lifelong MAPF problem by solving it as a whole~\cite{NguyenIJCAI17} or decomposing it into a sequence of MAPF instances and solving them on the fly~\cite{CapICAPS15,MaAAMAS17,WanICARCV18,LiuAAMAS19,GrenouilleauICAPS19,Li2020LifelongMP}.
We choose two state-of-the-art lifelong MAPF algorithms of two types as representative algorithms in our experiments, namely Dummy-Path Planning (DPP)~\cite{LiuAAMAS19} and Rolling-Horizon Collision Resolution (RHCR)~\cite{Li2020LifelongMP}. 

\paragraph{DPP} DPP plans paths for idle agents (i.e., agents that reach their current goal locations and need paths to their new goal locations) at every timestep. To ensure that there always exist paths for idle agents that do not collide with the paths of non-idle agents, it requires the layout to be well-formed and thus can be used only in the home-location scenario.

\paragraph{RHCR} RHCR replans paths for all agents at every $h$ timesteps, and the planned paths are provably collision-free for $w$ timesteps. It does not require the layout to be well-formed and thus can be used in both warehouse scenarios. %

\subsection{Existing Work on Layout Optimization}

To the best of our knowledge, no previous works have studied the warehouse layout optimization problem as described in \Cref{def:layout-opt} and explored the possibilities of having non-regularized warehouse layouts. 
Some related works include using queueing network models to estimate the throughput of a given warehouse layout~\cite{Wu2020ResearchOT,Lamballais2017EstimatingPI} and guide the design of warehouse layouts.
A recent work~\cite{Yang2021NonTraditionalLD} proposed using integer programming to optimize the locations of the workstations. However, it keeps the shelves fixed and evaluates the quality of the layouts by the average distance between the workstations and shelves, which does not reflect traffic congestion and is thus less accurate than our optimization objective throughput.

\subsection{Automatic Environment Generation and Quality-Diversity (QD) Optimization}

Automatic environment generation algorithms have been proposed in different fields such as procedural content generation~\cite{togelius2016procedural}, reinforcement learning~\cite{Risi2019IncreasingGI,Justesen2018IlluminatingGI}, and robotics~\cite{fontaine2021quality}. Among the environment generation techniques, QD algorithms~\cite{fontaine2021importance,fontaine2021quality,fontaine2020illuminating} are popular because they can find a diverse set of high-quality solutions with user-specified objectives and diversity measures.

MAP-Elites~\cite{Cully_2015_robot_animal,mouret2015illuminating} is a popular QD algorithm. It discretizes the measure space, which is defined by the user-specified diversity measure functions, and searches for the best solution in each discretized cell. The resulting best solutions in each cell become a set of diverse and high-quality solutions. Deep Surrogate Assisted MAP-Elites (DSA-ME)~\cite{Zhang2021DeepSA} improves MAP-Elites by using a surrogate model based on deep neural networks to speed up the solution evaluation process. DSAGE~\cite{Bhatt2022DeepSA} then extends DSA-ME to the environment generation problem for single-agent domains. %

\section{Layout Optimization Approach}

We first formulate the layout optimization problem as a QD problem and introduce our basic framework based on MAP-Elites, which searches over the tiles of the warehouse layout and uses a MILP solver to make the layout valid or well-formed. We then introduce our advanced framework based on DSAGE, which further introduces a surrogate model to improve optimization efficiency.

\subsection{MAP-Elites} \label{subsec:map-elites}

Since our warehouse layout is made of discrete tiles, we directly search over the possible tile combinations through a QD algorithm.
Specifically, we represent the storage area of a warehouse layout as a vector of discrete variables $\Vec{x} \in \mathbf{X}$, where each discrete variable $x_i\in\{black, blue, white\}$  corresponds to the tile type of the $i^{\text{th}}$ tile in the layout, and $\mathbf{X}$ is the space of all possible layouts.

To formulate layout optimization as a QD problem, we define an objective function to be maximized, $f: \mathbf{X} \rightarrow \mathbb{R}$, that runs a lifelong MAPF simulator on the input layout for $N_e$ times and returns the average throughput. 
We also define a measure function $\mathbf{m}: \mathbf{X} \rightarrow \mathbb{R}^m$ that quantifies characteristics of the layout to diversify. In our experiment, we select the number of connected shelf components and the average length of tasks in the layout as the diversity measures. We chose connected self components because shelves are often organized into ``islands'' of shelves in human-designed layouts.
We formalize connected shelf components as a set of shelves such that any pair of shelves in the set can be connected by a path through only the shelves in the set. In addition, we pick average task length because it is one of the factors that could affect throughput. We define the average length of tasks as the average distance between any pair of blue and pink tiles for the workstation scenario and any pair of two blue tiles for the home-location scenario.

We discretize the measure space evenly into $M$ cells and aim to find, for each cell $j$, the best layout $\Vec{x}_j$, called an \textit{elite layout}, %
that maximizes $f(\Vec{x}_j)$. We call this discretized measure space with elites an \textit{archive}. The bottom plot in \Cref{fig:exp-diversity} shows an example. %
Solving this problem generates a set of layouts with high throughput and different numbers of connected shelf components and average lengths of tasks.

\begin{figure*}[!ht]
    \centering
    \includegraphics[width=0.9\textwidth]{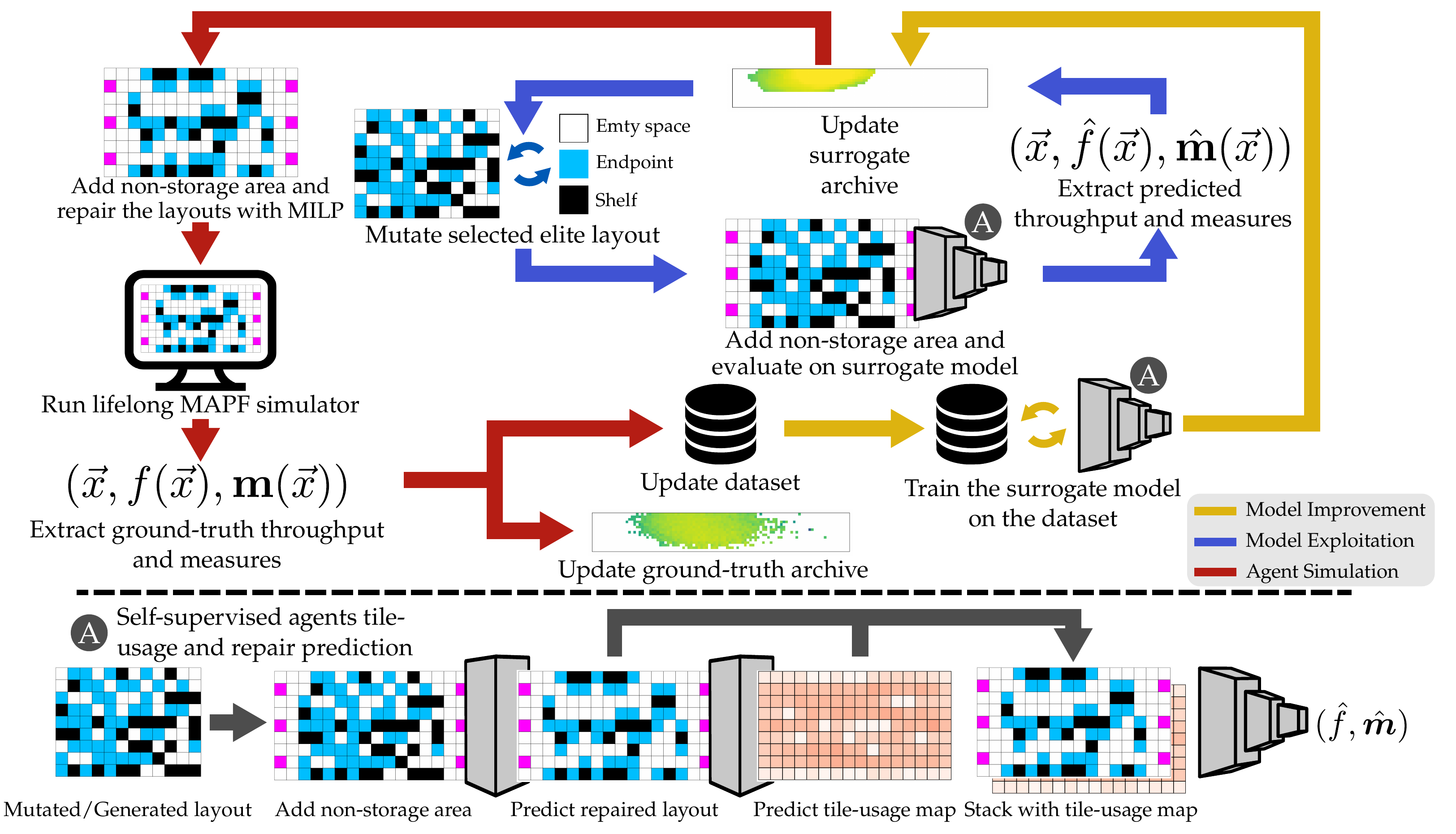}
    \caption{Overview of the warehouse layout optimization with extended DSAGE. We start by generating a randomly generated set of layouts and training the initial surrogate model. We then iterate over the model exploitation, agent simulation, and model improvement phases until the number of evaluations in the lifelong MAPF simulator reaches $N_{eval}$.}
    \label{fig:dsage}
\end{figure*}

\paragraph{Main Algorithm}
We maintain an archive and improve its elites through iterations.
In each iteration, we choose a batch of $b$ elite layouts from the archive uniformly with replacement, where the batch size $b$ is a hyperparameter. Inspired by previous work~\cite{fontaine:gecco19,fontaine2021importance}, we mutate each of the $b$ elite layouts by uniformly selecting $k$ tiles from each layout and changing the tile type of each of them to a random tile type. The variable $k$ is sampled from a geometric distribution with $P(X=k) = (1-p)^{k-1}p$ with $p=\frac{1}{2}$.
For the first iteration when the archive is empty, we just generate $b$ random layouts.
Then, we repair the layouts with a MILP solver to ensure that they are valid or well-formed (see details below). 
After getting the repaired layouts, we extract the average throughput and measures by running the lifelong MAPF simulator $N_e$ times, each runs for $T$ timesteps with $N_a$ agents. 
Finally, we add the evaluated layouts to the corresponding cells in the archive if their throughput is larger than that of the elite layouts in the same cells. We run MAP-Elites for $I$ iterations with batch size $b$, resulting in a total of $N_{eval} = b \times I$ evaluations, where each evaluation consists of $N_e$ simulations on the lifelong MAPF simulator. \Cref{fig:map-elites} in \Cref{appen:map-elites} gives an overview of MAP-Elites. %

\paragraph{MILP Repair}
After we generate or mutate a layout, we add the non-storage area back to the layout and repair it so that it becomes valid or well-formed.
We follow previous work~\cite{zhang:aiide2020,fontaine2021importance} and formulate the repair process as a Mixed Integer Linear Programming (MILP) problem. In particular, we minimize the hamming distance between the input unrepaired layout $\Vec{x}_{in}$ and the output repaired layout $\Vec{x}_{out}$ while asking $\Vec{x}_{out}$ to satisfy the following constraints:
(1) the tiles of $\Vec{x}_{out}$ in the non-storage area are kept unchanged, (2) $\Vec{x}_{out}$ is valid for the workstation scenario and well-formed for the home-location scenario, and (3) the number of shelves in $\Vec{x}_{out}$ is equal to a predefined number $N_s$, which ensures that $\Vec{x}_{out}$ has the desired storage capability. We constrain the number of shelves in the MILP repair step instead of in MAP-Elites so that the QD algorithm can search freely the solution space without being constrained to $N_s$.
The detailed formulation is shown in \Cref{appen:milp}.

\subsection{DSAGE}
Both the MILP solver and the lifelong MAPF simulator can be time-consuming for large layouts with large numbers of agents. For instance, in our experiments, the MILP solver can take as long as two minutes to repair one layout, and the lifelong MAPF simulator can take up to ten minutes to evaluate one layout. Therefore, we extend Deep Surrogate Assisted Generation of Environment (DSAGE)~\cite{Bhatt2022DeepSA} as an additional method to optimize warehouse layouts, which exploits an online trained deep surrogate model to guide MAP-Elites and improve its sample efficiency. 
To use DSAGE for layout optimization, we make the following extensions to DSAGE: (1) our model predicts the tile usage of hundreds of agents instead of a single agent, (2) the tile usage depends on the repaired layout, thus our surrogate model includes an additional component that predicts the output of the MILP solver.

\Cref{fig:dsage} gives an overview of our extended DSAGE framework, which consists of three phases (model exploitation, agent simulation, and model improvement), and a surrogate model. We start by generating a set of $n_{rand}$ random layouts and using them to train a surrogate model, which predicts the outputs of the MILP solver and the lifelong MAPF simulator for given unrepaired layouts. The model exploitation phase runs MAP-Elites with the surrogate model, instead of the MILP solver and the MAPF simulator, and gets a surrogate archive of (predicted) elite layouts. For every $n$ iterations of the model exploitation phase, we run the agent simulation phase, which repairs and evaluates the elite layouts from the surrogate archive using the MILP solver and the MAPF simulator and adds them to a ground-truth archive. The model improvement phase then saves all evaluated layouts in a dataset and uses it to improve the surrogate model.
We run this process iteratively until $N_{eval}$ layouts are evaluated on the lifelong MAPF simulator.
This way, we simultaneously generate a diverse set of high-quality layouts and train a deep surrogate model to predict the throughput and measures of given layouts. We show the details of the three phases below.

\paragraph{Model Exploitation} At this phase, we run MAP-Elites by replacing the MILP solver and the lifelong MAPF simulator with the surrogate model. That is, instead of computing the \textit{ground-truth} throughput $f(\Vec{x})$ and measures $\mathbf{m}(\Vec{x})$ using a repaired valid layout, we query the surrogate model for the \textit{predicted} throughput $\hat{f}(\Vec{x})$ and measures  $\hat{\mathbf{m}}(\Vec{x})$ using an unrepaired layout. Then the model exploitation phase efficiently builds a \textit{surrogate} archive, which contains the layouts that the surrogate model predicts to be high-performing and diverse. %

\paragraph{Agent Simulation} At this phase, we repair and evaluate the elite layouts from the surrogate archive to update the ground-truth archive. Since the surrogate archive sometimes contains a large number of elite layouts, we use the downsampling method from \cite{Bhatt2022DeepSA} to select a subset of elite layouts to be repaired and evaluated. Specifically, we uniformly divide the archive into sub-areas and uniformly select a layout in each sub-area. 
Besides updating the ground-truth archive, we also store the evaluated layouts in the dataset. 

\begin{figure}[!t]
    \centering
    \begin{subfigure}{0.23\textwidth}
        \centering
        \includegraphics[width=1\textwidth]{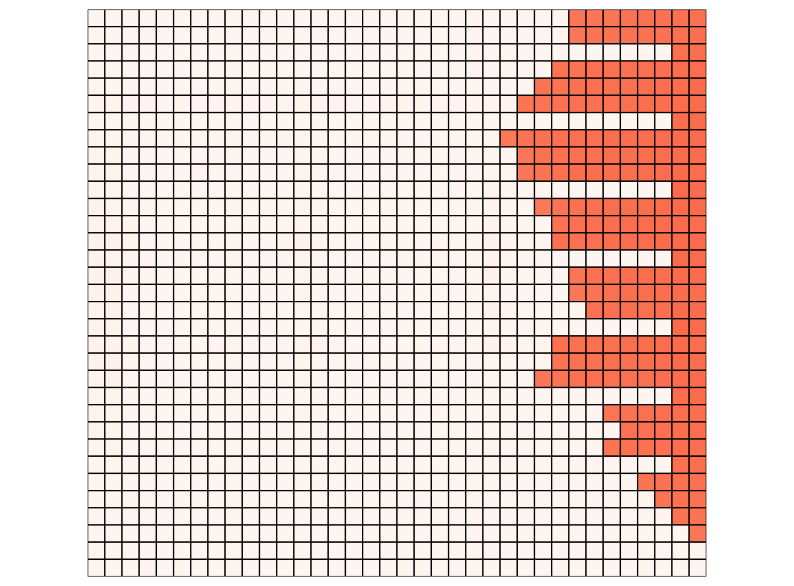}
        \caption{Human-designed layout}\label{fig:tile-usage-human}
    \end{subfigure}%
    \hfill
    \begin{subfigure}{0.23\textwidth}
        \centering
        \includegraphics[width=1\textwidth]{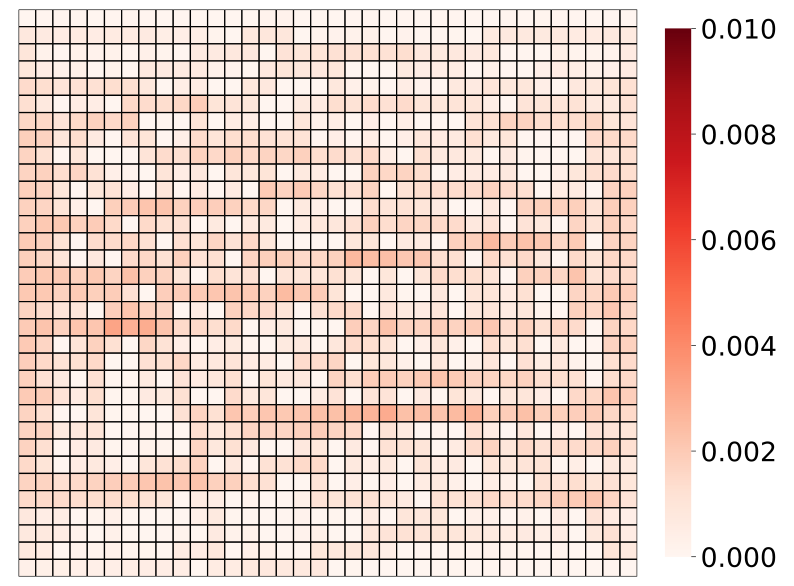}
        \caption{Optimized layout}\label{fig:tile-usage-optimized}
    \end{subfigure}%
    \caption{
    Tile-usage maps of the two layouts in \Cref{fig:front-fig:human,fig:front-fig:me-opt}. %
    }
    \label{fig:tile-usage}
\end{figure}

\paragraph{Model Improvement} At this phase, we fine-tune the surrogate model with the data in the dataset. %
We extend the surrogate model from previous work~\cite{Bhatt2022DeepSA} and use a three-stage deep convolutional neural network with a self-supervision procedure as our surrogate model. The architecture of the model is shown at the bottom of \Cref{fig:dsage}. It contains three sub-networks $\mathbf{s}_1$, $\mathbf{s}_2$, and $\mathbf{s}_3$. We first use $\mathbf{s}_1$ to predict a repaired layout given an input unrepaired layout. We then use $\mathbf{s}_2$ to predict a tile-usage map of the predicted repaired layout. A tile-usage map records the number of timesteps each tile is occupied by an agent in the lifelong MAPF simulation. We then normalize the tile-usage map so that the numbers sum up to 1, forming a distribution of the tile usage in the layout. We use the tile-usage map to self-supervise the training of the surrogate model because we observe that the distribution of the tile usage is correlated to the throughput of the layout. For example, \Cref{fig:tile-usage} shows the tile-usage maps of the two layouts in \Cref{fig:front-fig:human,fig:front-fig:me-opt} with 200 agents. \Cref{fig:tile-usage-human} indicates that the agents are congested on the right part of the warehouse, while \Cref{fig:tile-usage-optimized} has no sign of congestion. After predicting the tile-usage map, we concatenate the predicted tile-usage map with the predicted repaired layout and use $\mathbf{s}_3$ to predict the throughput and measures. We use a convolutional neural network with $48$ layers and $974,403$ parameters as the surrogate model, with its detailed architecture and training summarized in \Cref{appen:subsec:model}.

\section{Experimental Evaluation} \label{sec:exp}

In this section, we evaluate MAP-Elites and DSAGE and compare the optimized layouts with the human-designed ones. %

\subsection{Experiment Setup}

\begin{table}[!t]
    \centering
    \resizebox{\linewidth}{!}{
        \begin{tabular}{c|c|c|c|c|c|c|c|c}
        Setup & $S_{sa}$ & $|\mathbf{X}|$ & $S_f$ & $N_s$ & Scenario & MAPF & $N_a$ & $P_a$ \\
        \hline
        \multirow{2}{*}{1} & \multirow{2}{*}{$12 \times 9$} & \multirow{2}{*}{$3^{108}$} & \multirow{2}{*}{$20 \times 17$} & \multirow{2}{*}{$20$} &\multirow{2}{*}{Home location} & RHCR & \multirow{2}{*}{$88$} & \multirow{2}{*}{$28\%$} \\
         & &  &  &  & & DPP & &\\
        \hline
        \rule{0pt}{2ex}2 & $12 \times 9$ (small) & $3^{108}$ & $16 \times 9$ & $20$ & \multirow{3}{*}{Workstation} & \multirow{3}{*}{RHCR} & $60$ & $48\%$ \\
        3 & $12 \times 17$ (medium) & $3^{204}$ & $16 \times 17$ & $40$ & &  & $90$ & $39\%$ \\
        4 & $32 \times 33$ (large) & $3^{1056}$ & $36 \times 33$ & $240$ & &  & $200$ & $20\%$ 
        \end{tabular}
    }
    \caption{Summary of the experiment setup. $S_{sa}$ is the size of the storage area. $|\mathbf{X}|$ is the number of all possible layouts without constraints, $S_f$ is the full size of the layout, $N_s$ is the number of shelves, $N_a$ is the number of agents, and $P_a = \frac{N_a}{S_f -\ N_s}$ is the percentage of traversable tiles (i.e., non-black tiles) that are occupied by agents. 
    }
    \label{tab:search-space}
\end{table}

\Cref{tab:search-space} summarizes the experiment setup. 
Columns $2$-$6$ show the configurations related to the layouts. We use one map size for the home-location scenario and three map sizes for the workstation scenario (to show how our methods scale with the size of the maps). %
Notably, setup $4$ has the same storage area size as a commonly used human-designed layout in previous works~\cite{Li2020LifelongMP}. The number of shelves $N_s$ shown in column $5$ is chosen to be the same as those in the human-designed layouts so that the optimized layouts have the same storage capacity as the human-designed layouts.

Column $7$ shows the lifelong MAPF simulator with RHCR and DPP. Following design recommendation of previous work~\cite{Li2020LifelongMP}, we use PBS~\cite{MaAAAI19} as the MAPF solver with $w = 10$ and $h = 5$ in RHCR, and use prioritized planning~\cite{Erdmann1987PP} as the MAPF solver in DPP. Both of them use SIPP~\cite{PhillipsICRA11} as the single-agent solver. In the workstation scenario, the initial locations of the agents are uniformly chosen at random from non-shelf tiles, and the task assigner alternates every agent's goal location between a workstation and an endpoint, both chosen uniformly at random. In the home-location scenario, the initial locations of the agents are uniformly chosen at random from the home locations, and the task assigner assigns uniformly random goal locations from the endpoints.
Since DPP can only work on well-formed layouts, we use it only for the home-location scenario.

Columns $8$ and $9$ show the number of agents $N_a$ and the agent density $P_a$. 
For the home-location scenario, we choose $N_a=88$ because we have only $88$ home locations.
For the workstation scenario, we run the MAPF simulator on the human-designed layouts with an increasing number of agents and choose $N_a$ when we observe a significant drop in throughput. 

For both MAP-Elites and DSAGE, we set $b=50$, $N_{eval} = 10,000$, $T = 1,000$, and $N_e=5$. 
For DSAGE, we further set $n_{rand} = 500$ and $n = 10,000$ for setup $1$ and $2$ and $n = 50,000$ for setup $3$ and $4$. 
We stop the lifelong MAPF simulation early (i.e., before timestep $T$) if (unrecoverable) \emph{congestion} occurs, which is defined as the case where more than half of the agents take wait actions at a particular timestep. %
This is because (1) we penalize the layouts that run into congestion and (2) we empirically observe that congestion can quickly decrease the number of finished tasks, and keeping running the simulation with congestion seldom increases the throughput. We summarize other hyperparameters and compute resources in \Cref{appen:subsec:archive}.

\begin{figure*}[!t]
    \centering
    \includegraphics[width=1\textwidth]{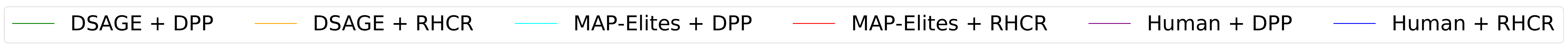}\\
     \begin{subfigure}{0.24\textwidth}
        \centering
        \includegraphics[width=1\textwidth]{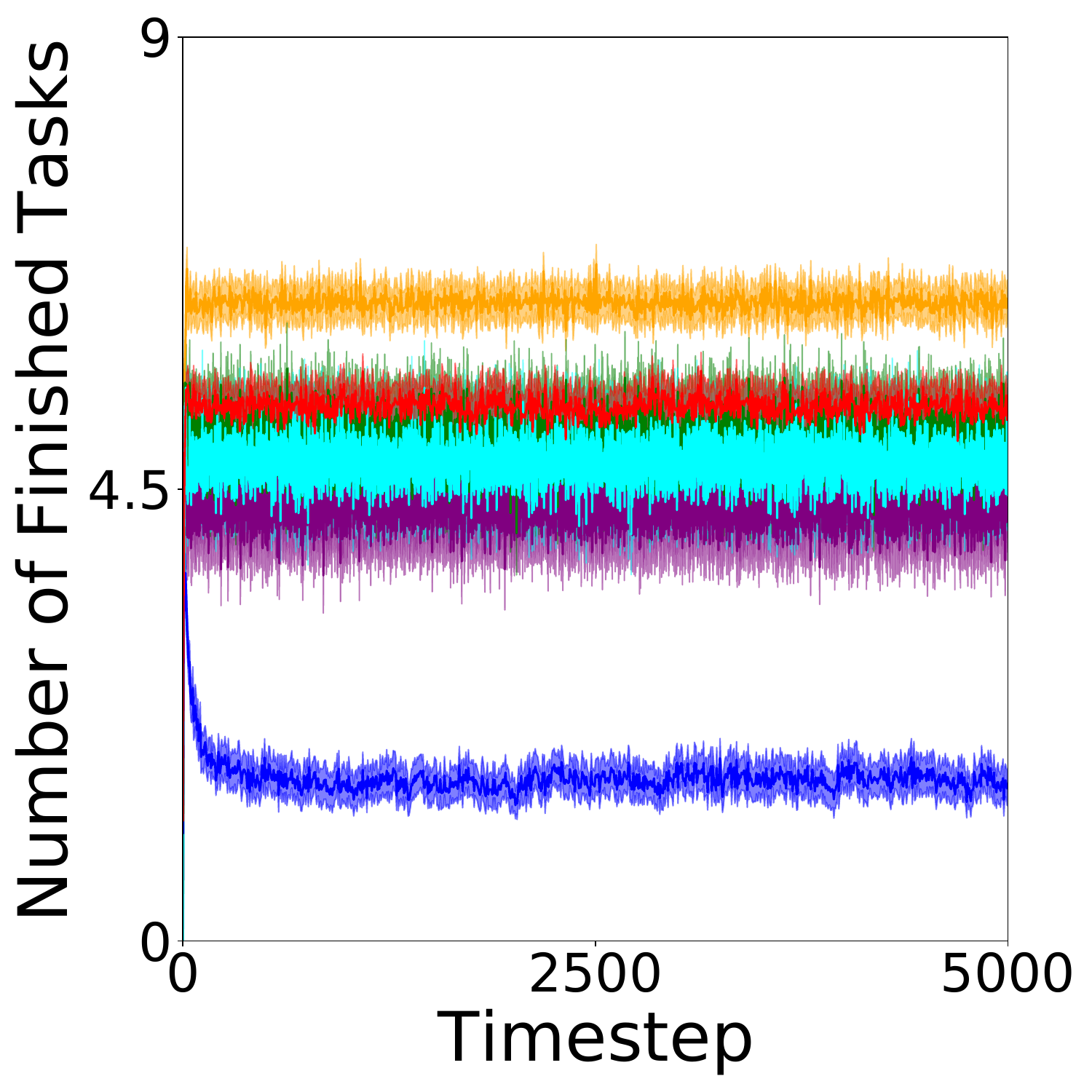}
        \caption{Setup $1$}
        \label{fig:thr-time-small-r}
    \end{subfigure}%
    \hspace{0.005\textwidth}
    \begin{subfigure}{0.24\textwidth}
        \centering
        \includegraphics[width=1\textwidth]{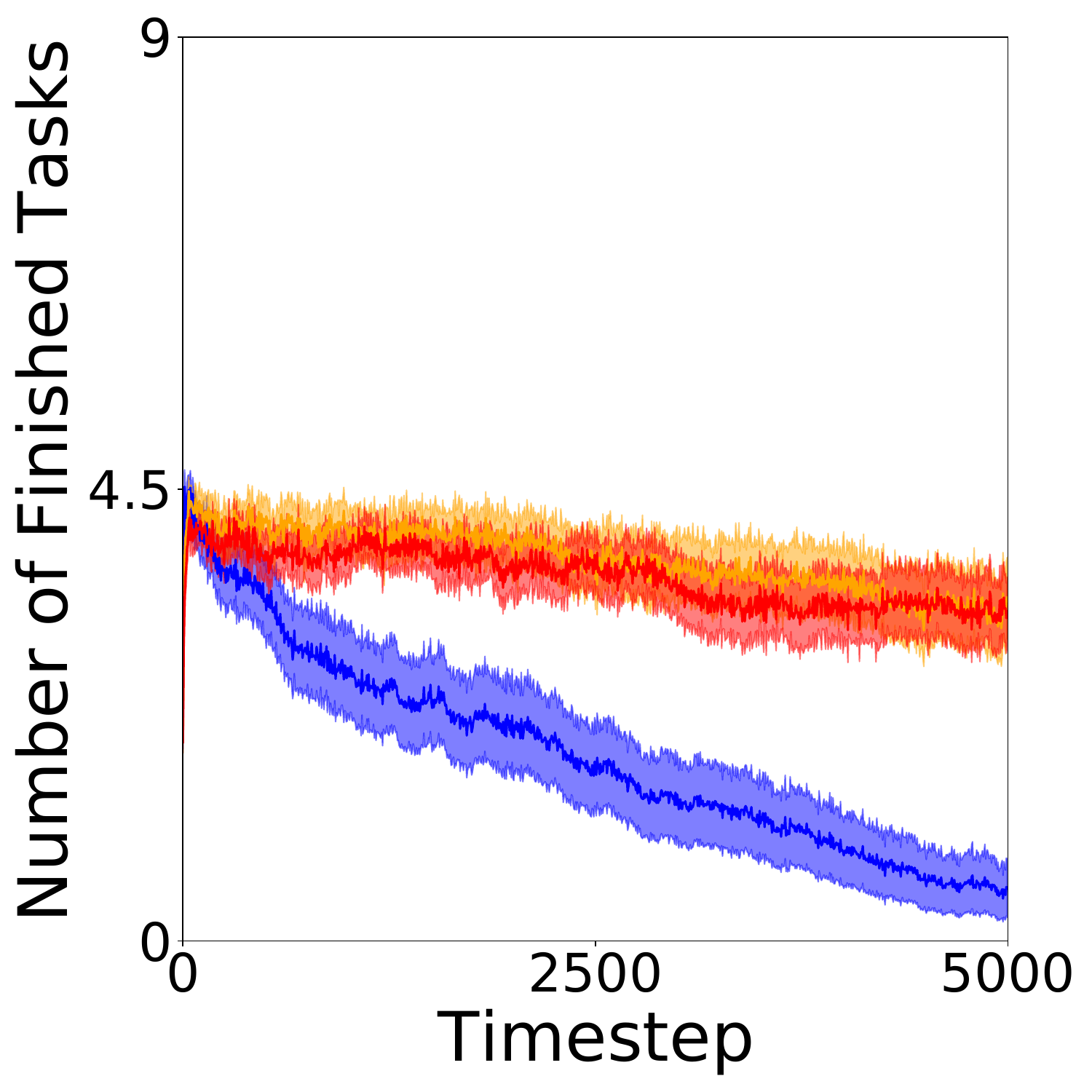}
        \caption{Setup $2$}
        \label{fig:thr-time-small-w}
    \end{subfigure}%
    \hspace{0.005\textwidth}
    \begin{subfigure}{0.24\textwidth}
        \centering
        \includegraphics[width=1\textwidth]{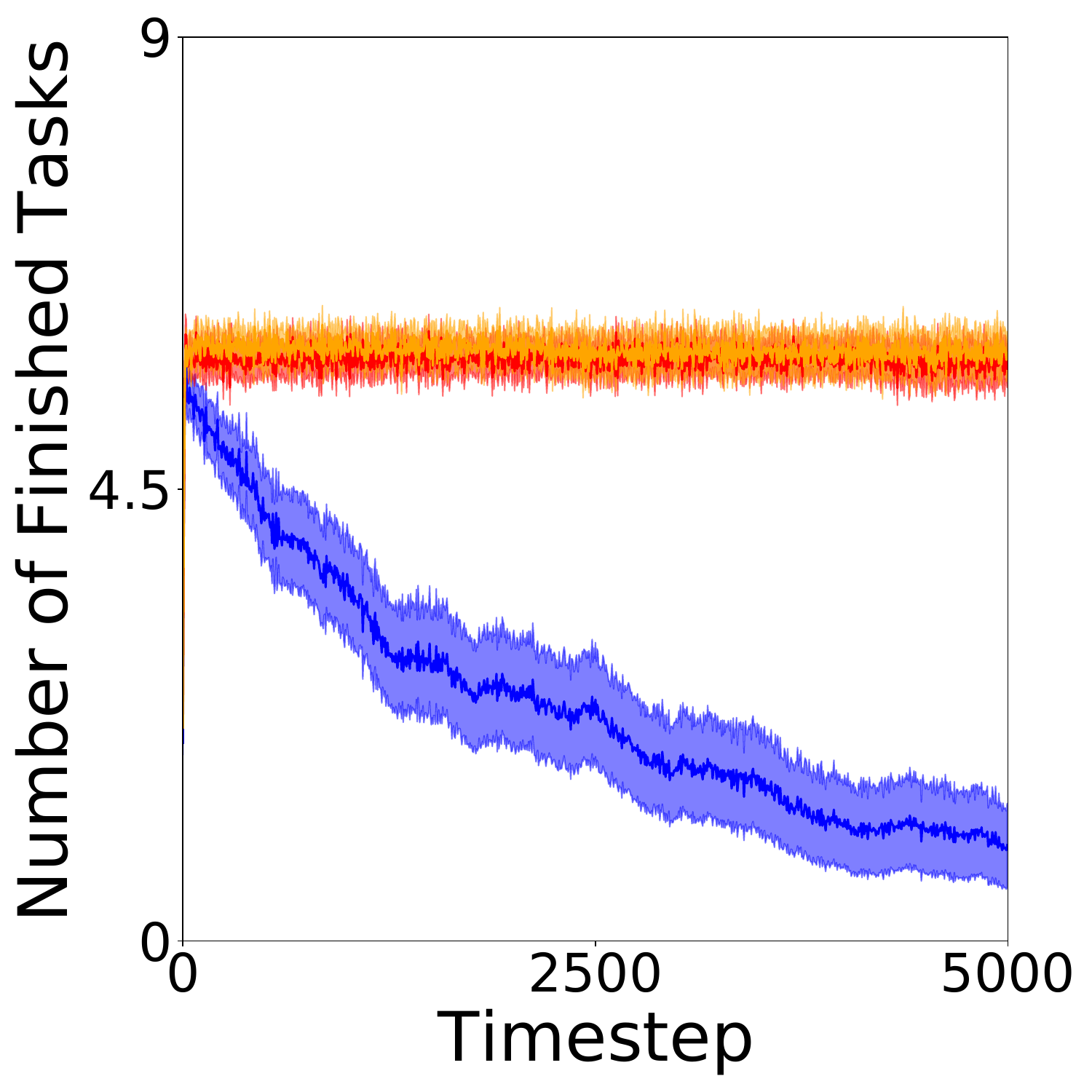}
        \caption{Setup $3$}
        \label{fig:thr-time-medium-w}
    \end{subfigure}%
    \hspace{0.005\textwidth}
    \begin{subfigure}{0.24\textwidth}
        \centering
        \includegraphics[width=1\textwidth]{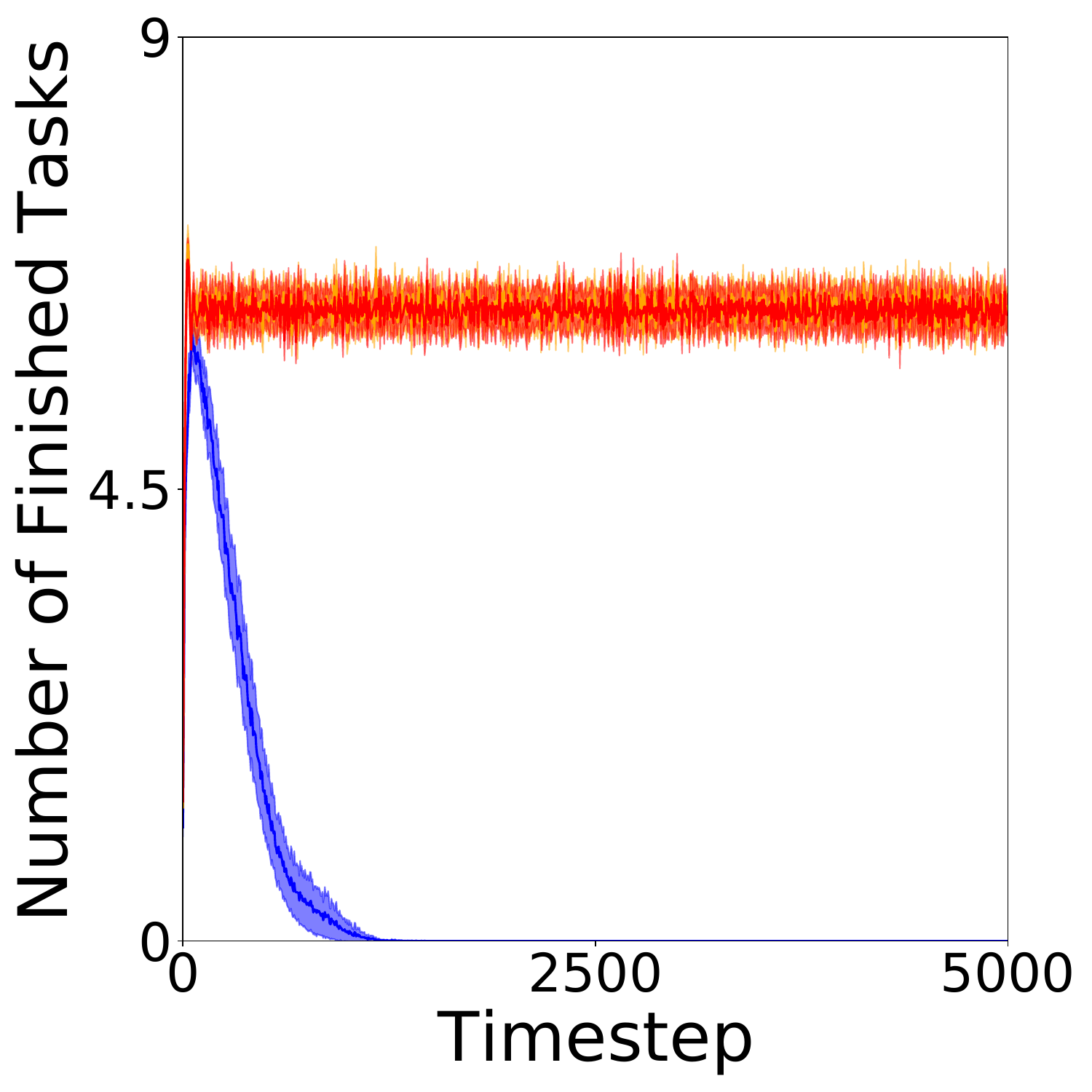}
        \caption{Setup $4$}
        \label{fig:thr-time-large-w}
    \end{subfigure}\\
    \begin{subfigure}{0.24\textwidth}
        \centering
        \includegraphics[width=1\textwidth]{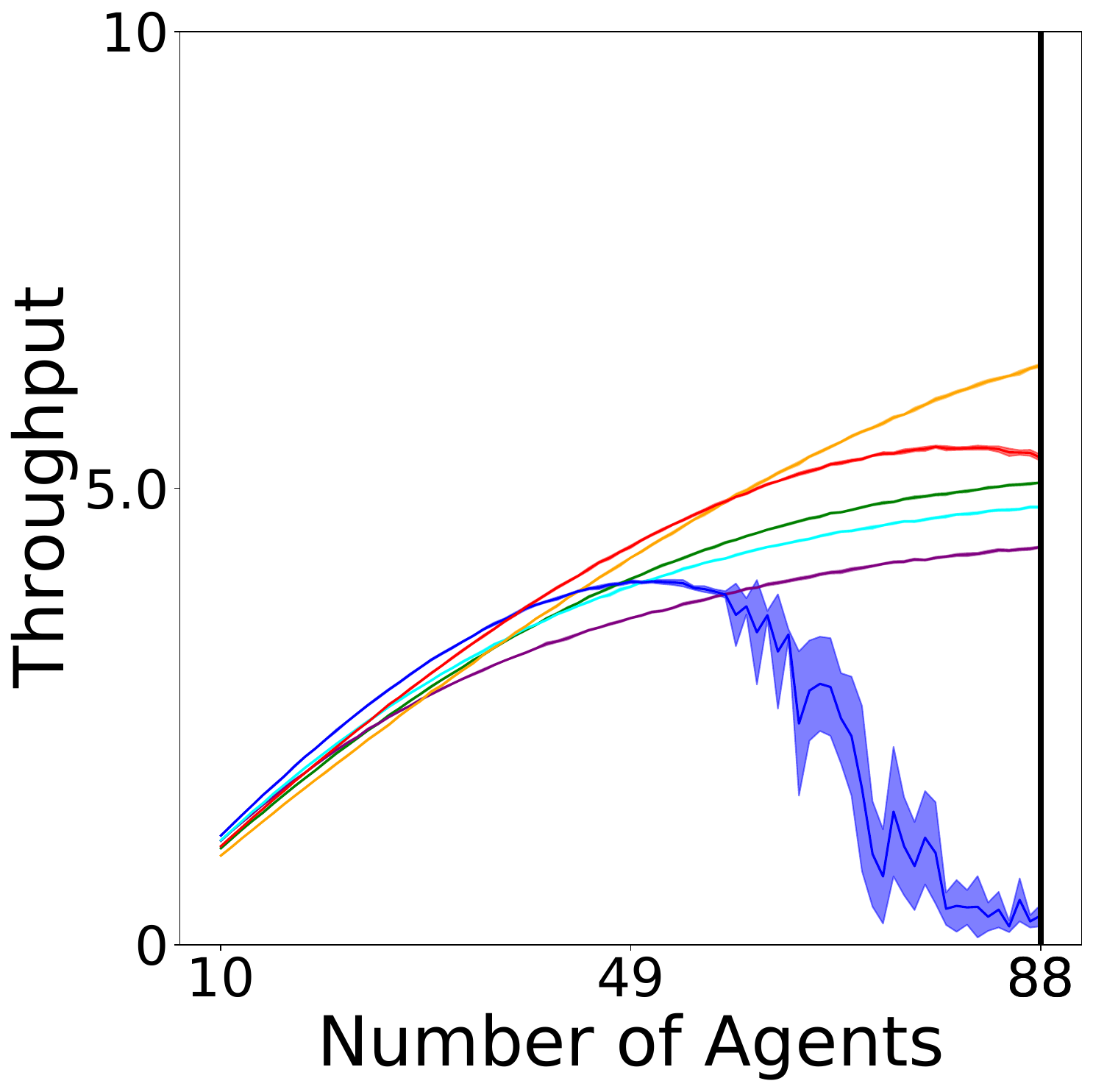}
        \caption{Setup $1$}
        \label{fig:agent-num-small-r}
    \end{subfigure}%
    \hspace{0.005\textwidth}
    \begin{subfigure}{0.24\textwidth}
        \centering
        \includegraphics[width=1\textwidth]{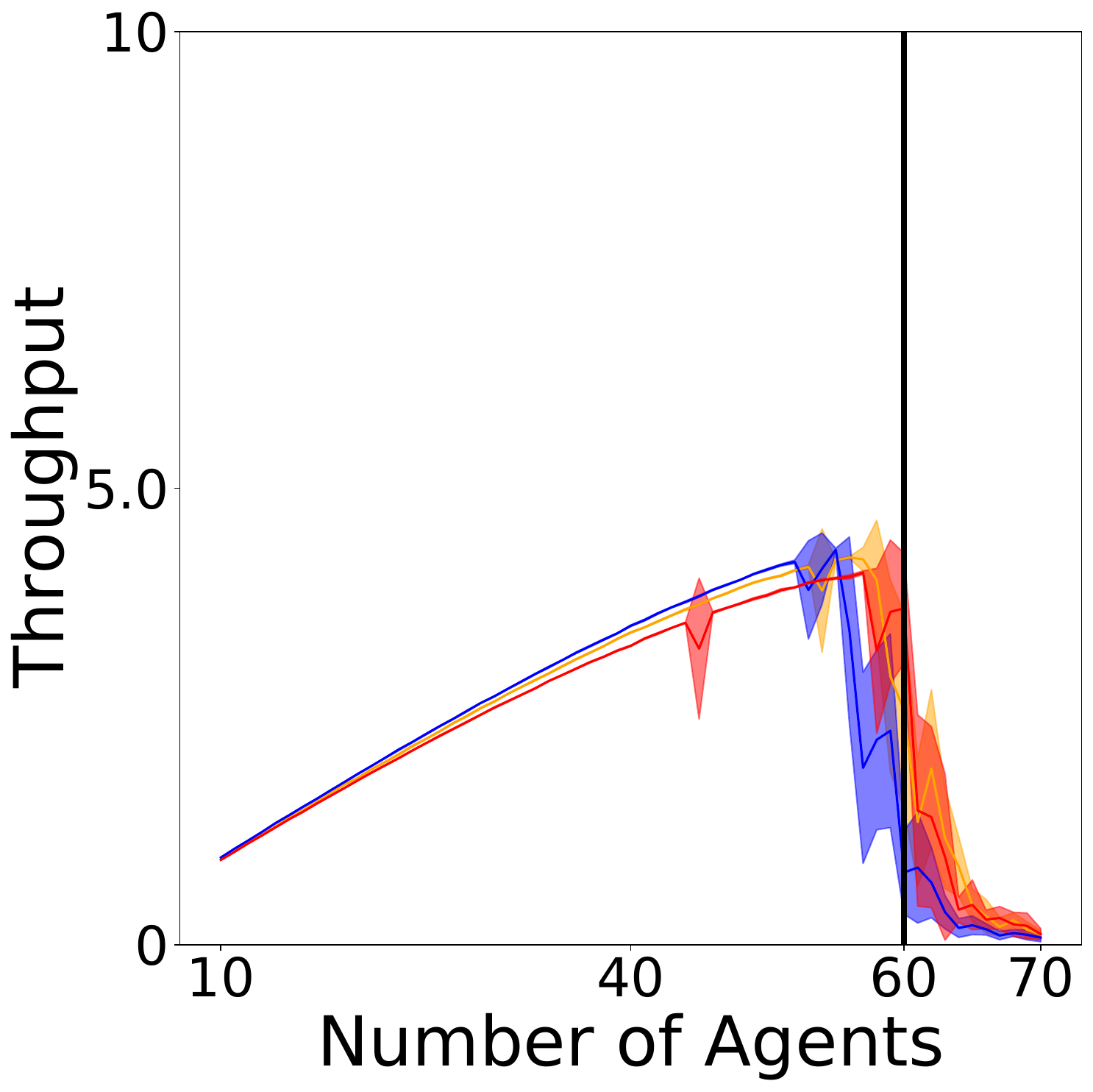}
        \caption{Setup $2$}
        \label{fig:agent-num-small-w}
    \end{subfigure}%
    \hspace{0.005\textwidth}
    \begin{subfigure}{0.24\textwidth}
        \centering
        \includegraphics[width=1\textwidth]{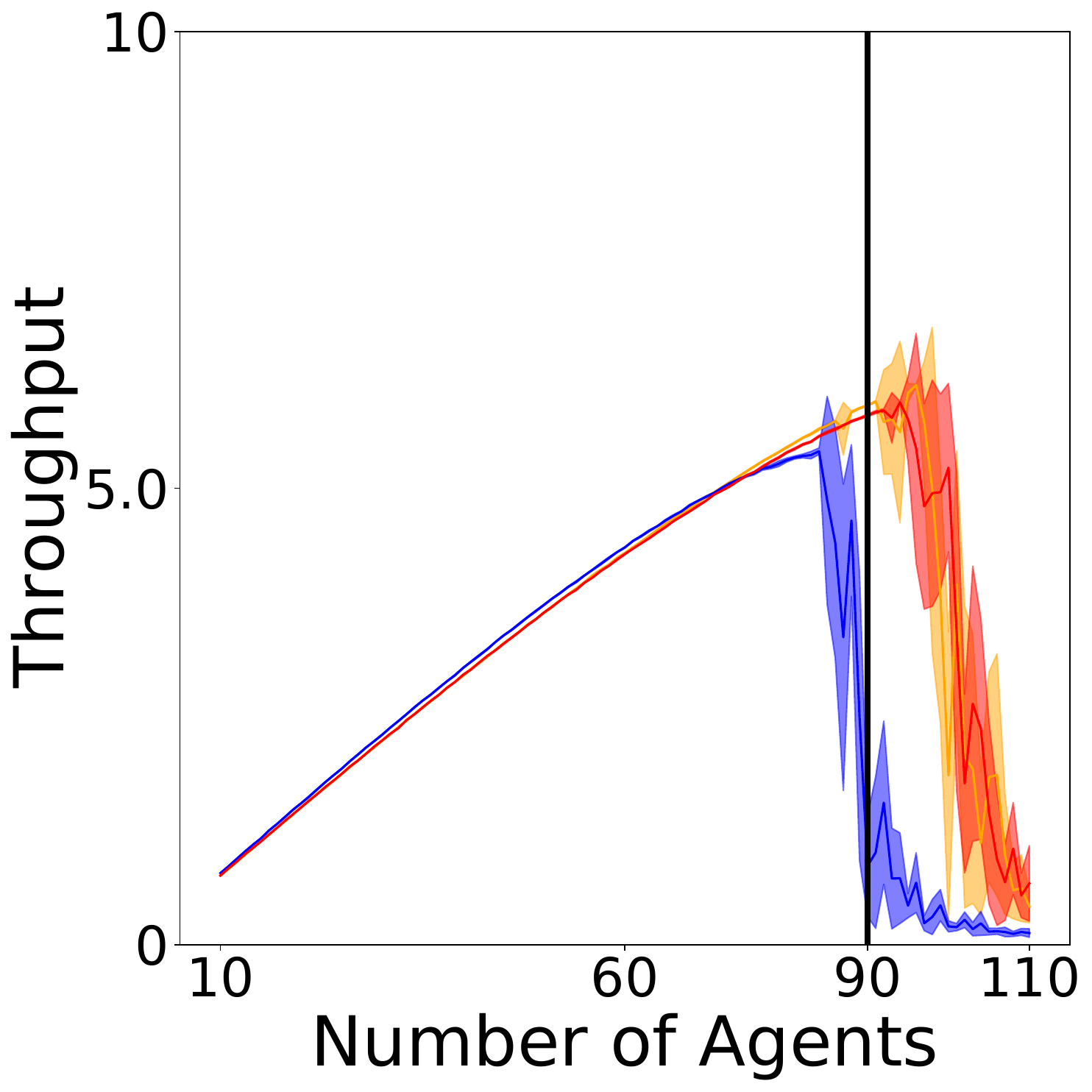}
        \caption{Setup $3$}
        \label{fig:agent-num-medium-w}
    \end{subfigure}%
    \hspace{0.005\textwidth}
    \begin{subfigure}{0.24\textwidth}
        \centering
        \includegraphics[width=1\textwidth]{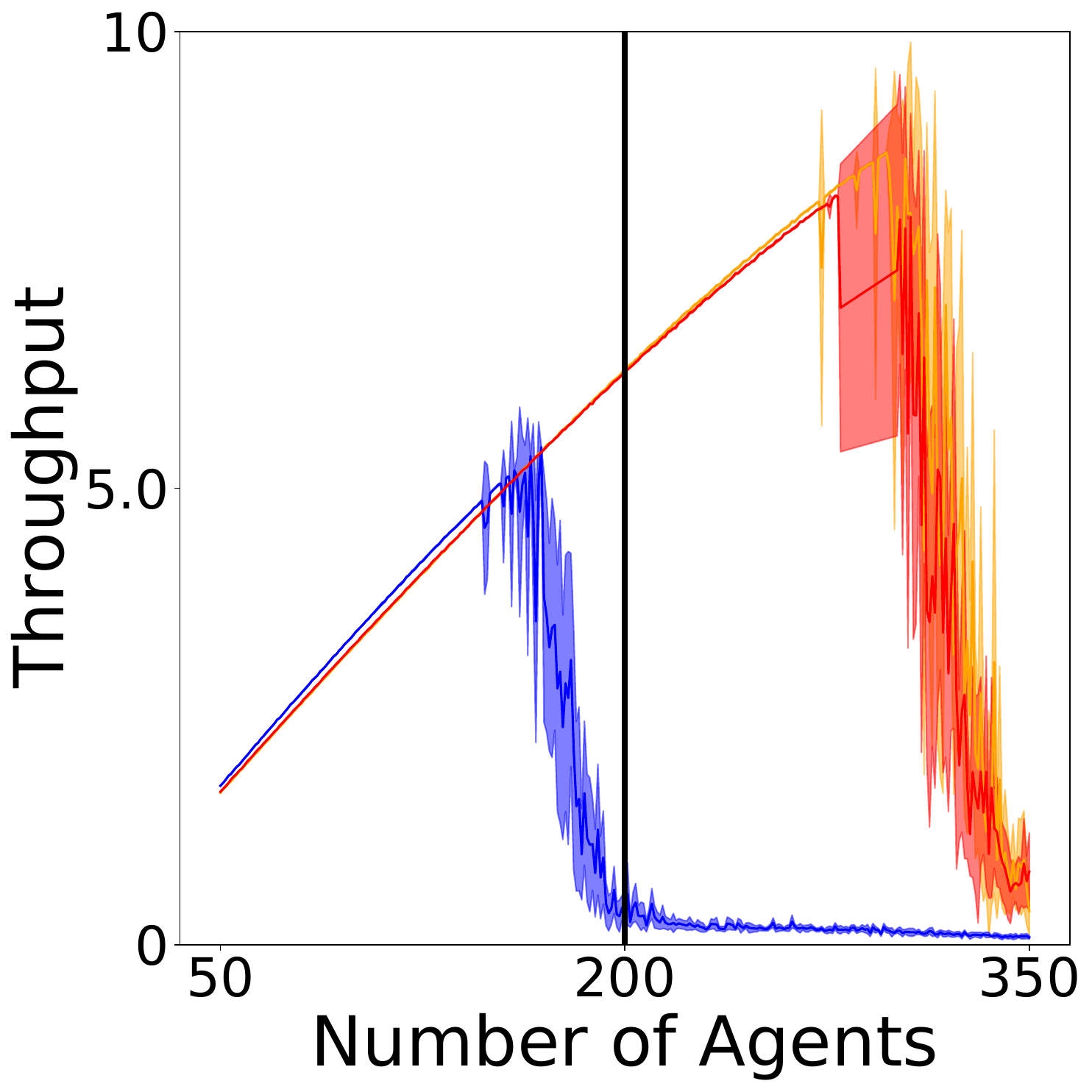}
        \caption{Setup $4$}
        \label{fig:agent-num-large-w}
    \end{subfigure}
    \caption{Number of finished tasks per timestep with $N_a$ agents and throughput with different numbers of agents. The layouts are optimized with $N_a$ agents, where $N_a$ is indicated by the vertical lines in \Cref{fig:agent-num-small-r,fig:agent-num-small-w,fig:agent-num-medium-w,fig:agent-num-large-w}.}
    \label{fig:major-result}
\end{figure*}

\begin{table}[!t]
    \centering
    \resizebox{\linewidth}{!}{
        \begin{tabular}{c|c|c|c|c}
        Setup               & QD + MAPF Algo              &  SR                & Throughput                        & CPU Runtime \\
        \hline
        \multirow{6}{*}{1}  &\textbf{DSAGE + RHCR}        &  $\textbf{100\%}$  & $\textbf{6.34} \pm \textbf{0.01}$ & $174.62 \pm 1.97$  \\
                            & MAP-Elites + RHCR           &  $\textbf{100\%}$  & $5.33 \pm 0.00$                   & $382.80 \pm 2.75$  \\
                            & Human + RHCR                &  $0\%$             & N/A                               & N/A  \\
                            & DSAGE + DPP                 &  $\textbf{100\%}$  & $5.06 \pm 0.00$                   & $\textbf{46.42} \pm \textbf{0.99}$ \\
                            & MAP-Elites + DPP            &  $\textbf{100\%}$  & $4.79 \pm 0.00$                   & $47.14 \pm 1.41$ \\
                            & Human + DPP                 &  $\textbf{100\%}$  & $4.36 \pm 0.00$                   & $55.71 \pm 0.96$ \\
        \hline
        \multirow{3}{*}{2}  & \textbf{DSAGE + RHCR}       &  $30\%$           & $\textbf{4.28} \pm \textbf{0.02}$  & $\textbf{73.13} \pm \textbf{1.61}$ \\
                            & MAP-Elites + RHCR           &  $\textbf{80\%}$  & $4.08 \pm 0.02$                    & $83.36 \pm 1.36$ \\
                            & Human + RHCR                &  $0\%$            & N/A                                & N/A \\
        \hline
        \multirow{3}{*}{3}  &\textbf{DSAGE + RHCR}        &  $\textbf{100\%}$ & $\textbf{5.91} \pm \textbf{0.00}$  & $\textbf{93.66} \pm \textbf{0.42}$ \\
                            & MAP-Elites + RHCR           &  $\textbf{100\%}$ & $5.80 \pm 0.00$                    & $103.55 \pm 0.47$ \\
                            & Human + RHCR                &  $0\%$            & N/A                                & N/A \\
        \hline
        \multirow{3}{*}{4}  &  \textbf{DSAGE + RHCR}      &  $\textbf{100\%}$ & $\textbf{6.30} \pm \textbf{0.00}$  & $\textbf{131.63} \pm \textbf{3.32}$  \\
                            & MAP-Elites + RHCR           &  $\textbf{100\%}$ & $6.27 \pm 0.00$                    & $152.13 \pm 1.16$ \\
                            & Human + RHCR                &  $0\%$            & N/A                                & N/A

        \end{tabular}
    }
    \caption{Success rates, throughput, and CPU runtimes. %
    \textit{SR} refers to the success rate, which is the percentage of simulations that end without congestion. \textit{CPU runtime} refers to the runtime, in seconds, of the $5000$-timestep simulation. We only measure the throughput and CPU runtime of successful simulations.}
    \label{tab:numerical-result}
\end{table}

To report robust results, we run the lifelong MAPF simulator on every human-designed or optimized layout that we evaluate in this section with $T_{eval}=5,000$ timesteps for, unless explicitly stated otherwise, $10$ times, using machines specified in \Cref{append:subsec:compute}. We report both average results and their distributions (in the format of $x \pm y$ for numerical results with standard deviations and solid lines with shared areas for graphical results with $95$\% confidence intervals).

\subsection{Results} \label{subsec:result}

\subsubsection{Comparing Optimized and Human-Designed Layouts}
We take the best layout from the archive for each setup and compare them in \Cref{tab:numerical-result}.
The results show that our optimized layouts have significantly better throughput and less congestion than the human-designed layouts in all cases. For example, we scale RHCR from complete failure to complete success for all setups except for setup $2$. %
Moreover, we find that, for the same lifelong MAPF algorithm, a layout with better throughput often leads to a smaller CPU runtime of the lifelong MAPF simulator. Thus, we also reduce the runtime of DPP in the setup $1$.

To further understand the gap of the success rates and throughput among different layouts, we plot the simulation process in \Cref{fig:thr-time-small-r,fig:thr-time-small-w,fig:thr-time-medium-w,fig:thr-time-large-w}. Here, we do not stop the simulation even if they run into congestion and report results over $100$ runs (instead of $10$ runs). %
As shown in the figures, the human-design layouts often encounter severe unrecoverable congestion, leading to a significant drop in the number of finished tasks over time. In contrast, our optimized layouts are stable in most cases. The only exception is setup $2$, where the performance of our optimized layouts also decreases over time, though with a smaller decreasing speed than that of the human-design layout. This is because setup 2 has the highest agent density (see $P_a$ in \Cref{tab:search-space}), which makes it hard to find layouts that do not cause congestion.

\subsubsection{Generalization and Scalability}
Although we run MAP-Elites and DSAGE with a fixed number of agents $N_a$, the optimized layouts can also be applied to other numbers of agents. 
\Cref{fig:agent-num-small-r,fig:agent-num-small-w,fig:agent-num-medium-w,fig:agent-num-large-w} shows the throughput with different numbers of agents on layouts that are optimized with $N_a$ agents. In comparison to the human-designed layouts, our optimized layouts achieve similar throughput for small numbers of agents and significantly better throughput for large numbers of agents in setup $1$. As for the workstation scenario (i.e., setups $2$-$4$), the trend is similar, and the larger the layout, the bigger the improvement. For instance, we double the scalability of RHCR from fewer than $150$ agents to more than $300$ agents in setup $4$.

\begin{figure}[!t]
    \centering
    \includegraphics[width=0.4\textwidth]{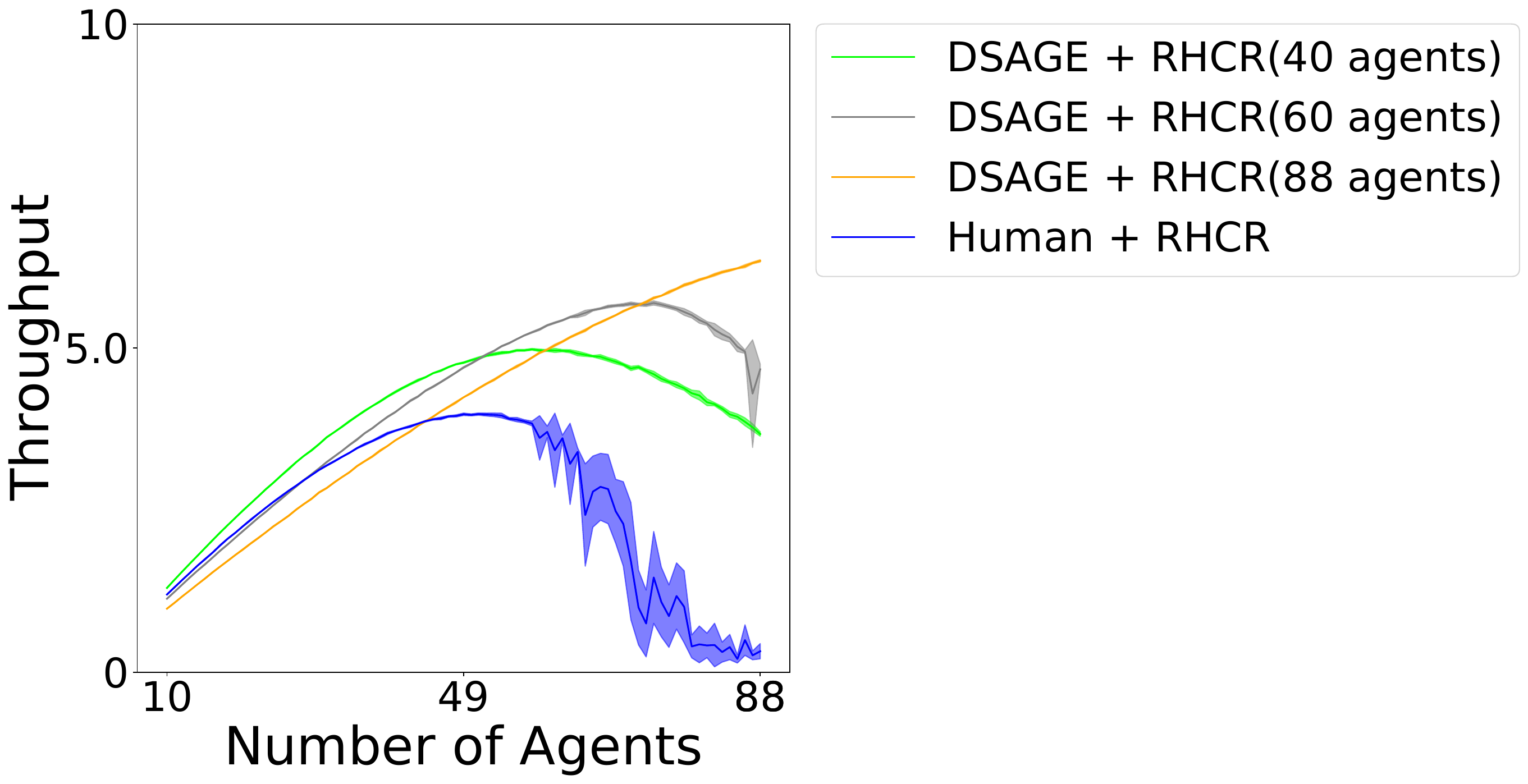}
    \caption{Throughput for layouts optimized with different numbers of agents for setup 1.}
    \label{fig:less-agents}
\end{figure}

One reason why we do not improve the throughput for small numbers of agents is that our layouts are optimized using large numbers of agents. To justify this, we optimize our layouts using three different numbers of agents in setup $1$ and report the results in \Cref{fig:less-agents}. As shown, when we use a small number of agents to optimize the layout, the resulting layout leads to better throughput for small numbers of agents.

We also observe an interesting trade-off between RHCR and DPP. Previous work~\cite{Li2020LifelongMP} shows that RHCR leads to better throughput than DPP, which matches our results for small numbers of agents. But for large numbers of agents, we observe that RHCR is more prone to congestion and thus performs worse than DPP. Nevertheless, our optimized layouts improve the performance of both of them.

\subsubsection{Diversity in Generated Layouts}

\begin{figure}[!t]
    \centering
    \includegraphics[width=0.47\textwidth]{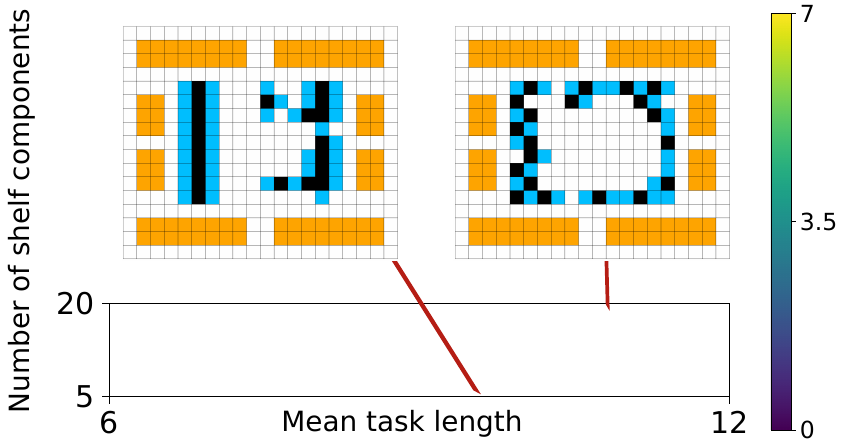}
    \caption{Example optimized layouts with different numbers of connected shelf components from the archive. Layouts are optimized with DSAGE and RHCR with $N_a=88$ agents. The layouts on the left and right achieved throughput of $6.03$ and $4.80$, respectively.}
    \label{fig:exp-diversity}
\end{figure}

A major benefit of using QD algorithms to optimize warehouse layouts is to get a diverse set of high-throughput layouts with user-defined diversity measures. \Cref{fig:exp-diversity} shows two example optimized layouts with different numbers of connected shelf components. The layout on the left clustered more shelves together, forming 5 components, whereas the layout on the right scatters the shelves in the storage area, forming 19 components. Notably, both layouts achieve higher throughput than the human-designed layout. In addition, the optimized layouts allocate the shelves and endpoints on the borders of the storage area. The geometric structure allows the MAPF algorithms to improve throughput.

We observe that DSAGE is more capable of generating diverse layouts in the archive than MAP-Elites. For example, in setup $4$, although there is no significant difference between the throughput of the best optimized layouts generated by MAP-Elites and DSAGE with $200$ agents, DSAGE generates $1,317$ elite layouts in the archive, and MAP-Elites generates only $1,039$. As a result, DSAGE generates layouts with numbers of connected shelf components ranging from $146$ to $218$, while MAP-Elites cannot generate layouts with fewer than $153$ or more than $207$ connected shelf components. We show more results on DSAGE and MAP-Elites in \Cref{appen:archive}.

\section{Conclusion}

We present the first layout optimization approach for automated warehouses that explores novel layouts with non-regularized patterns and explicitly optimizes throughput. We extend DSAGE from single-agent domains to the warehouse domain with as many as $200$ agents. We also incorporate MILP to DSAGE and extend the surrogate model to predict repaired layouts.
Our result shows that the optimized layouts outperform commonly used human-designed layouts in throughput and CPU runtime. Our methods work with different warehouse scenarios, different MAPF algorithms, and layouts of different sizes. 
In addition, our methods provide diverse layouts with respect to user-specified diversity measures. %

Our work has many future directions. First, we can incorporate the optimization of the non-storage area into our method because previous work~\cite{Lamballais2017EstimatingPI} has demonstrated the impact of workstations on throughput.
Second, our optimization method only works for small $2$D warehouses represented as a $4$-neighbor grid. Future works can focus on generating larger $2$D or $3$D warehouses.
Third, we can improve our method to optimize layouts with a larger agent density ($P_a$). 
Overall, we are excited about the first proposed method for layout optimization and are looking forward to adapting our approach to more complex warehouse domains.

\section*{Acknowledgments}

This work used Bridge-$2$ at Pittsburgh Supercomputing Center (PSC) through allocation CIS$220115$ from the Advanced Cyberinfrastructure Coordination Ecosystem: Services \& Support (ACCESS) program, which is supported by National Science Foundation grants \#$2138259$, \#$2138286$, \#$2138307$, \#$2137603$, and \#$2138296$. In addition, this work was supported by the NSF CAREER Award (\#$2145077$) and the CMU Manufacturing Futures Institute, made possible by the Richard King Mellon Foundation.

\bibliographystyle{named}
\bibliography{ijcai23}

\begin{figure*}
    \centering
    \includegraphics[width=.8\textwidth]{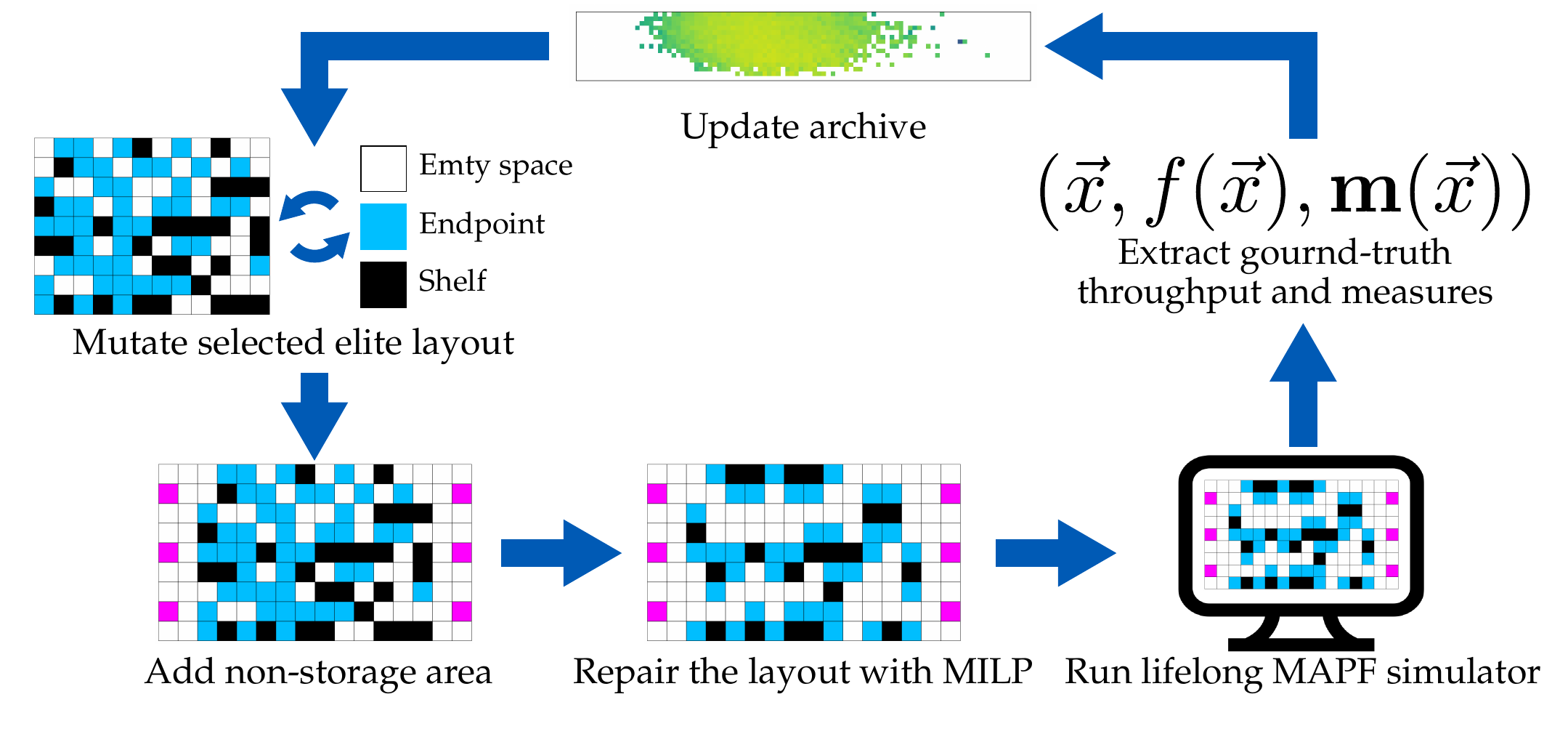}
    \caption{Overview of the warehouse layout optimization with MAP-Elites. We start by generating an initial archive with random layouts. We then iteratively select random elite layouts from the archive, mutate the selected elite layouts, add non-storage area, and repair the layout with MILP. Finally, we extract the throughput and measures by running the lifelong MAPF simulator and update the archive with the evaluated layouts. We iterate over the process until the number of evaluations in the lifelong MAPF simulator reaches $N_{eval}$.}
    \label{fig:map-elites}
\end{figure*}

\newpage
\appendix

\section{Human-designed Layouts} \label{appen:human-layout}

\begin{figure}[!ht]
    \centering
    \begin{subfigure}[t]{0.24\textwidth}
        \centering
        \includegraphics[width=1\textwidth]{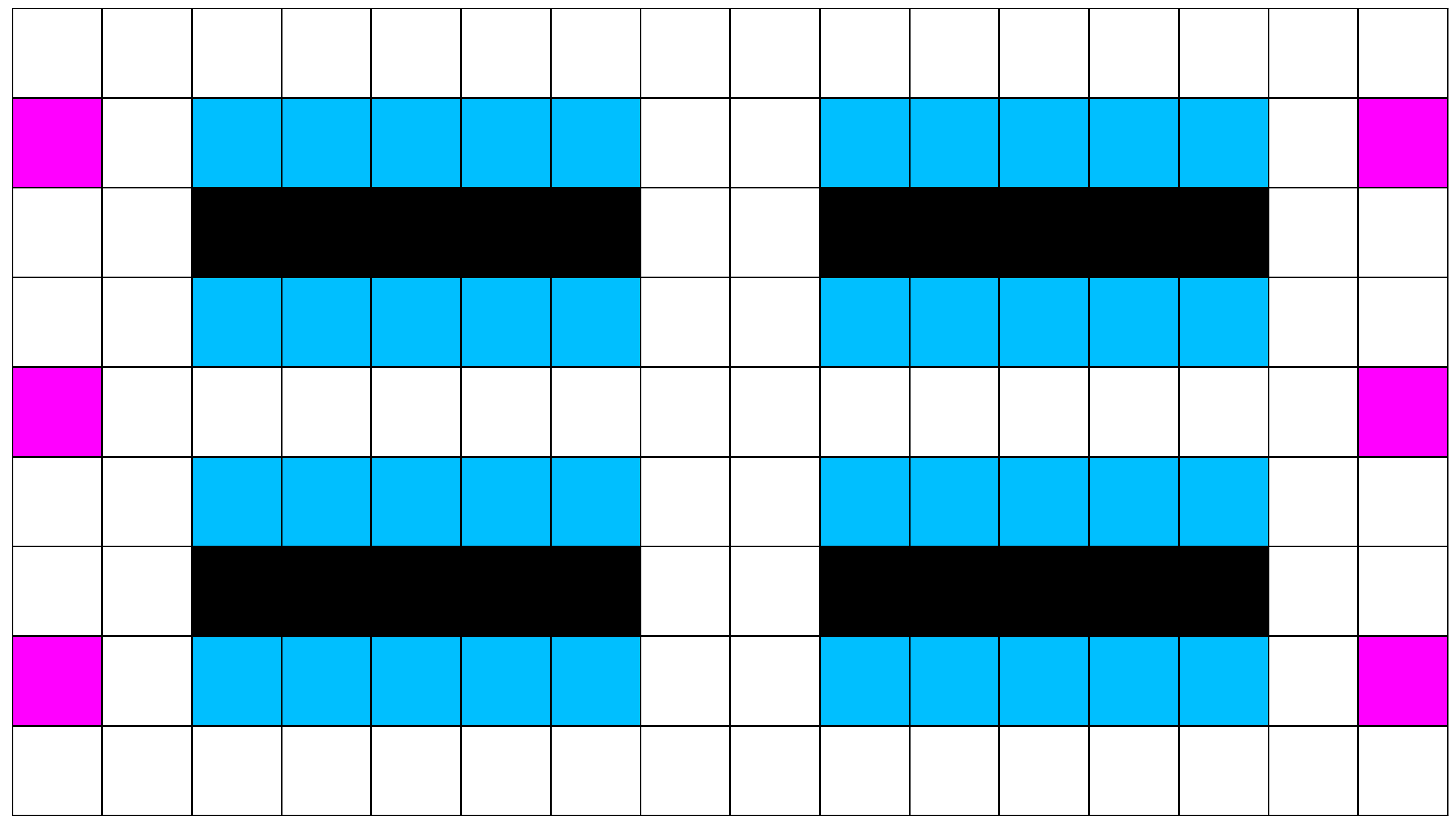}
        \caption{$16 \times 9$ workstation scenario}
        \label{fig:human-small-w}
    \end{subfigure}%
    \hfill
    \begin{subfigure}[t]{0.24\textwidth}
        \centering
        \includegraphics[width=1\textwidth]{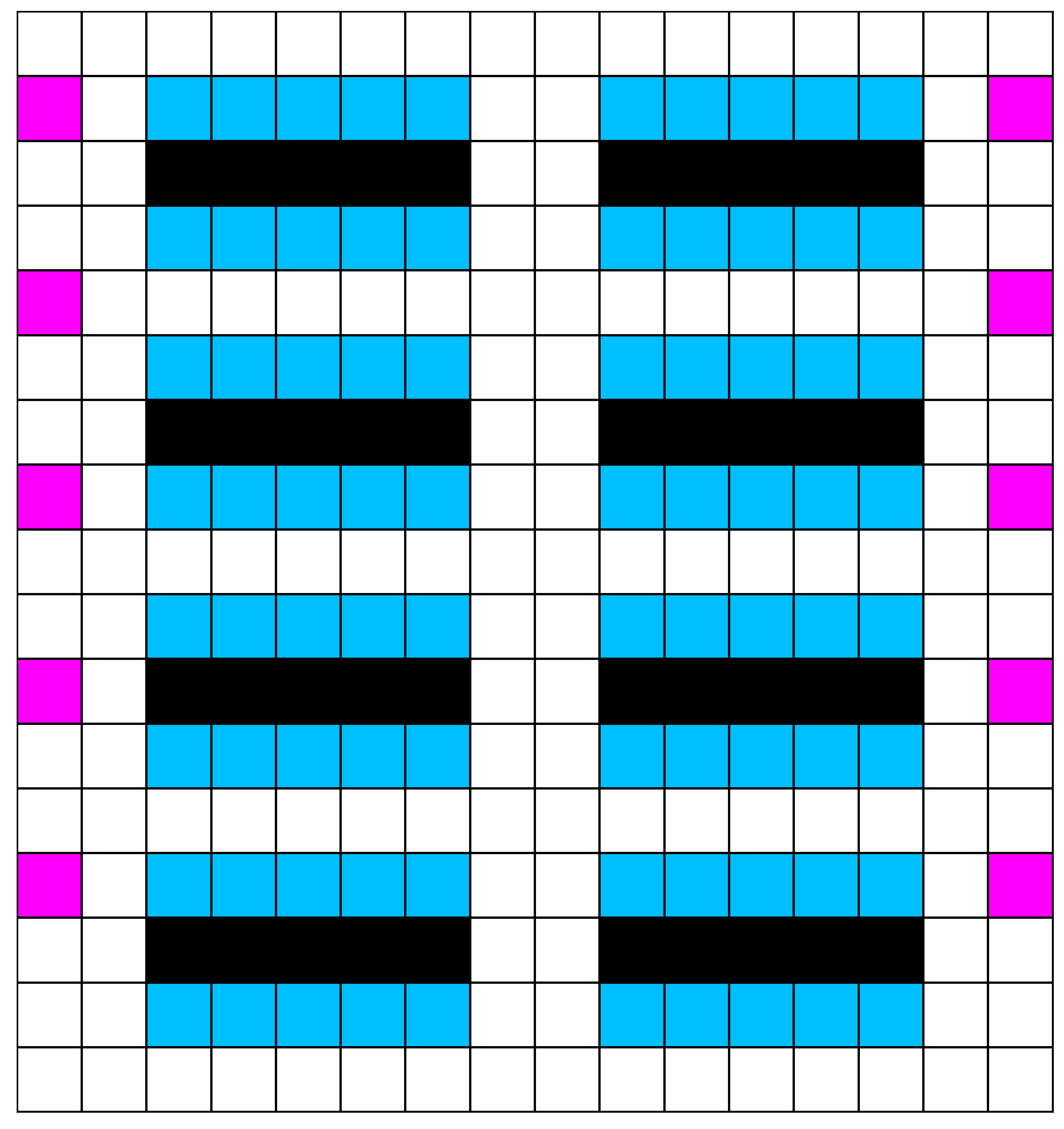}
        \caption{$16 \times 17$ workstation scenario}
        \label{fig:human-medium-w}
    \end{subfigure}\\
    \begin{subfigure}[t]{0.24\textwidth}
        \centering
        \includegraphics[width=1\textwidth]{maps/kiva_large_w_mode.png}
        \caption{$36 \times 33$ workstation scenario}
        \label{fig:human-large-w}
    \end{subfigure}%
    \hfill
    \begin{subfigure}[t]{0.24\textwidth}
        \centering
        \includegraphics[width=1\textwidth]{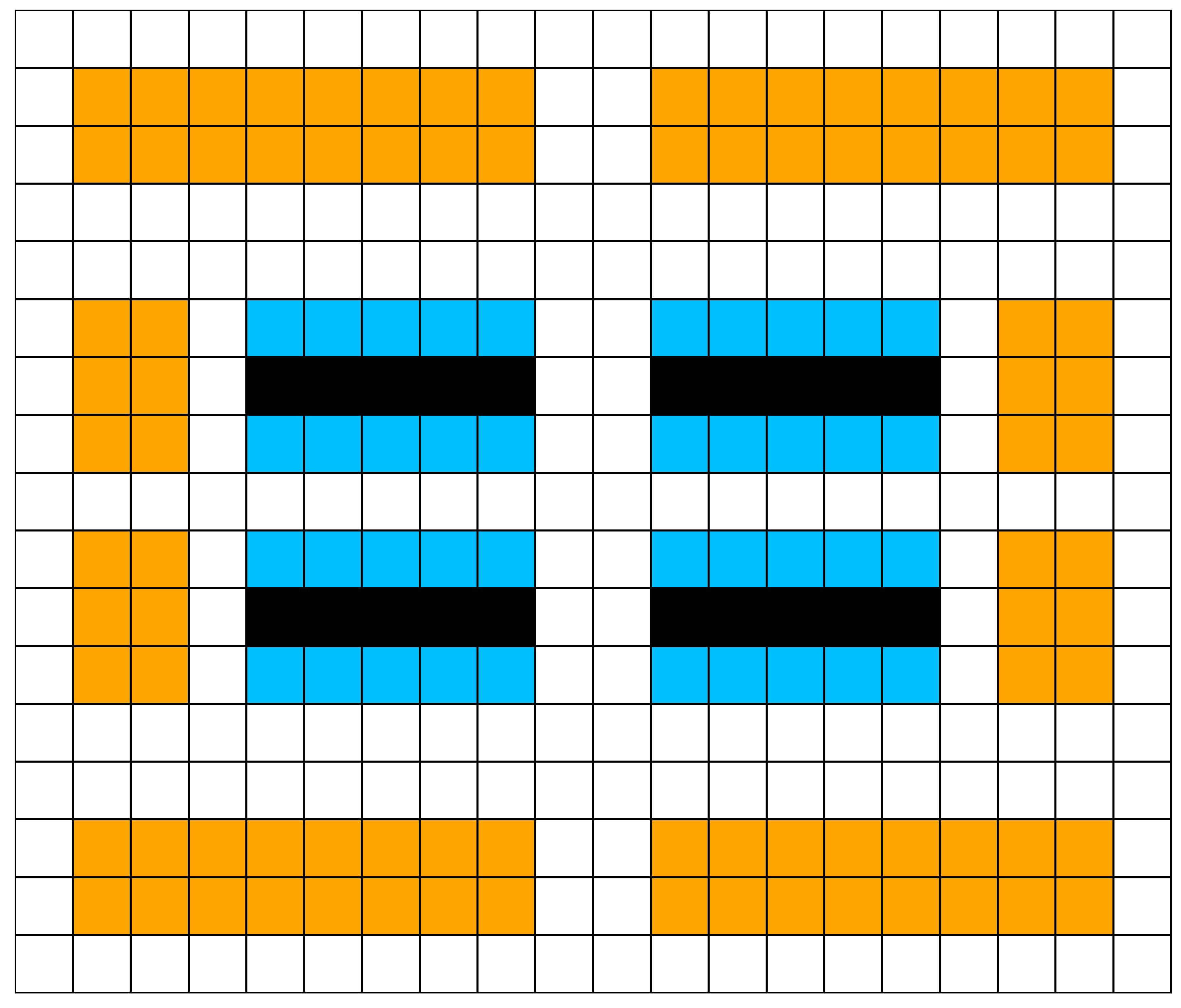}
        \caption{$20 \times 17$ home-location scenario}
        \label{fig:human-small-r}
    \end{subfigure}%
    \caption{Human-designed layouts}
    \label{fig:human-design}
\end{figure}

\Cref{fig:human-design} shows the human-designed layouts that we use as the baseline layouts. The storage area of $36 \times 33$ workstation scenario in \Cref{fig:human-large-w} is the same as the layouts used in previous works~\cite{Li2020LifelongMP,LiuAAMAS19}. It clusters the shelves into rows of $10$ and puts endpoints on the rows above and below the shelves. It then repeats the same pattern to design the entire layout. Following this regularized pattern, we design the other three human-designed layouts in \Cref{fig:human-small-w,fig:human-medium-w,fig:human-small-r}. We then append the non-storage area on the borders of the storage area.

\section{MAP-Elites} \label{appen:map-elites}

\Cref{fig:map-elites} shows the overview of warehouse layout optimization with MAP-Elites discussed in \Cref{subsec:map-elites}.

\section{Mixed Integer Linear Programming Formulation} \label{appen:milp}

We adapt the MILP formulation from previous work~\cite{fontaine2021importance,zhang:aiide2020} that repairs tile-based game levels for \textit{The Legend of Zelda} and \textit{Overcooked} to the warehouse domain by changing tile types and adding constraints specific to the warehouse domain.

We represent the warehouse layout (including both storage and non-storage areas) as a directed graph~\cite{togelius2016procedural} $G(V, E)$ where $V$ is a set of vertices and $E$ is a set of edges. Each vertex $v \in V$ represents a tile in the layout that contains exactly one tile type. For any two neighboring vertices $u$ and $v$, we build directed edges $(u,v)$ and $(v,u)$. The graph $G$ has the same edges as the warehouse layout which allows the agents to move to adjacent vertices on the left, right, top, or bottom if the adjacent vertex is not a shelf. Let $O$ be the set of tile types, which includes the home location $h$, the workstation $w$, the endpoint $e$, the shelf $s$, and the empty space $p$. 

To repair a given unrepaired warehouse layout, we solve for a matching between vertex set $V$ and tile type set $O$ such that the hamming distance between the unrepaired and repaired layout is minimized. Therefore, for each vertex $v \in V$ and each tile type $x \in O$, we define a binary variable $x_v \in \{0,1\}$ to indicate whether the tile type of vertex $v$ in the repaired layout is $x$ and a binary constant $x_v^0 \in \{0,1\}$ to indicate whether the tile type of vertex $v$ in the unrepaired layout is $x$. %
Our objective of minimizing the hamming distance between the repaired and unrepaired layouts can be then written as
\begin{equation}\label{eqn:obj}
    \min \sum_{v \in V, x \in O} |x_v - x_v^0|.
\end{equation}

Our first constraint is to ensure that each vertex corresponds to exactly one tile type.
\begin{equation}\label{eqn:uniqueness}
    \forall v \in V,\ h_v + w_v + e_v + s_v + p_v = 1.
\end{equation}%
We set $h_v = 0$ for the workstation scenario because we do not have home locations in this scenario. Similarly, we set $w_v = 0$ for the home-location scenario because we do not have workstations in this scenario.

Then, we ask the repaired layout to meet certain constraints, including (1) being valid for the workstation scenario and well-formed for the home-location scenario, (2) containing exactly $N_s$ shelves, and (3) keeping the tiles of non-storage area unchanged. %
We explain the details of these three constraints in the following.

\subsection{Non-Storage Area Constraints} \label{subsec:milp:non-storage}

First, we add constraints to fix the tiles in the non-storage area. We add the same non-storage area as the human-designed layouts shown in \Cref{fig:human-design}.
For workstation scenario, we define $W$ to be the set of vertices in the non-storage area. We add the following constraints to guarantee that they do not change:

\begin{equation}
    \forall v \in W,
    \begin{cases}
      w_v = 1 & \text{if } w_v^0 = 1\\
      p_v = 1 & \text{if } p_v^0 = 1
    \end{cases}   
\end{equation}

Similarly, for the home-location scenario, 
we define $H$ to be the set of vertices in the non-storage area. We add the following constraints to guarantee that they do not change:

\begin{table}[!t]
    \centering
        \begin{tabular}{c|c|c}
        Setup & $N_w$ & $N_h$\\
        \hline
        1  & N/A & $88$ \\
        2  & $6$ & N/A \\
        3  & $10$ & N/A \\
        4  & $22$ & N/A \\
        \end{tabular}
    \caption{Number of workstations and home-locations in each setup.}
    \label{tab:num-w-r}
\end{table}

\begin{equation}
    \forall v \in H,
    \begin{cases}
      h_v = 1 & \text{if } h_v^0 = 1\\
      p_v = 1 & \text{if } p_v^0 = 1
    \end{cases}   
\end{equation}

Furthermore, we constrain the total number of workstations and home locations in the layout so that the MILP solver does not put more workstations or home locations in the storage area. Specifically, given the fixed number of workstations $N_w$ and fixed number of home locations $N_h$ in \Cref{tab:num-w-r}, we have the following constraints:

\begin{equation}
    \sum_{v \in V} w_v = N_w
\end{equation}

\begin{equation}
    \sum_{v \in V} h_v = N_h.
\end{equation}

\subsection{Validity and Well-form Constraints}

We constrain the repaired layouts to be valid for the workstation scenario and well-formed for the home-location scenario. We first introduce how we constrain the layouts to be valid and then extend the constraints to the well-formed case.

\subsubsection{Shelf-Endpoint Adjacency Constraints}

We require tiles of shelves and endpoints to be adjacent in a valid layout. Specifically, we add the following constraints:

\begin{equation}
    \forall v \in V,\ \sum_{u:(u, v) \in E} s_u \geq e_v \label{eq:sum-s-e}
\end{equation}

\begin{equation}
    \forall v \in V,\ \frac{1}{2} \sum_{u:(u, v) \in E} e_u \geq s_v. \label{eq:sum-e-s}
\end{equation}%
\Cref{eq:sum-s-e} guarantees that each endpoint is adjacent to at least one shelf. Mathematically, if no adjacent vertices of a vertex $v$ are shelves, the left side of the inequality is $0$, %
indicating that $v$ cannot be an endpoint. Otherwise, if there is at least one shelf among the adjacent vertices of vertex $v$, %
the left side of the inequality is no smaller than $1$, indicating that $v$ can be an endpoint.
\Cref{eq:sum-e-s} is analogical and ensures that a vertex $v$ can be a shelf only if at least two adjacent vertices are endpoints.
We want at least two endpoints around each shelf so that more agents can interact with the same shelf in parallel.

\subsubsection{Reachability Constraints} \label{subsec:reachability}

We define a reachability constraint that requires all non-black tiles to be connected through non-black tiles. Obviously, a layout that satisfies the reachability constraint is valid. We use this reachability constraint, a more restrictive requirement than the requirement of valid layouts, because it ensures that every non-black tile can be used by agents and thus maximizes space utilization. For example, suppose a layout has an empty space surrounded by four shelves. Although it does not make the layout invalid, that empty space cannot be utilized by any agent.

\begin{figure*}[!t]
    \centering
    \includegraphics[width=1\textwidth]{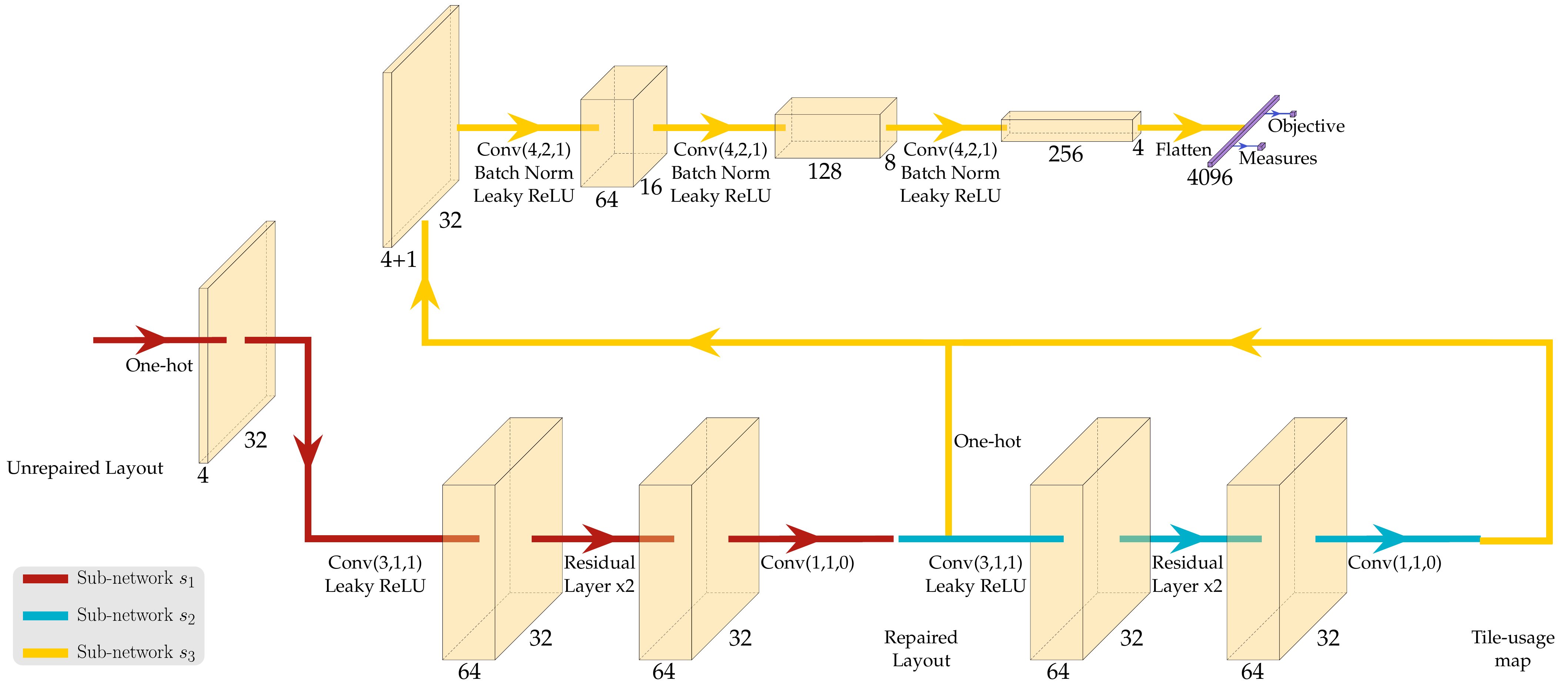}
    \caption{Architecture of the surrogate model for setup 2 in \Cref{sec:exp}. The model predicts the repaired layout from the unrepaired layout, which is then used to predict the tile-usage map. The one-hot encoding of the predicted repaired layout is concatenated with the predicted tile-usage map to predict the objective and measures.}
    \label{fig::surrogate-model-arch}
\end{figure*}

Following previous work~\cite{Goldberg2014flow}, we reduce the problem of making the layout satisfy the reachability constraint to a flow problem. We first arbitrarily pick a vertex with the workstation or home-location tile type as the source vertex. 
In our implementation, we introduce a new tile type called dummy source $d$ and a new tile-type binary variable $d_v \in \{0,1\}$ for each vertex $v \in V$. We force it to be $1$ for the picked source vertex and $0$ otherwise and add $d_v$ to the left-hand side of \Cref{eqn:uniqueness}. Since the tiles in the non-storage area do not change, the selection of this dummy vertex does not change either during the optimization. We will change the tile type back to the workstation or home location after repairing the layout. We define $T = O \setminus \{d, s\}$ as the set of sink tile types %
and $B = \{s\}$ as the set of blocking tile types. We call a vertex of a sink tile type a sink vertex.
Our goal is to find a flow in the graph such that there is a path from the source vertex to every sink vertex while not passing through any vertex of any blocking tile type. If such a flow exists, it indicates that the workstation or home location that the dummy source vertex corresponds to is connected to every non-black tile in the layout (through paths with only non-black tiles), which further indicates that the layout satisfies the reachability constraint. 

Formally, we define a non-negative continuous flow variable $f(u, v) \in \mathbb{R}_{\geq 0}$ for each edge $e = (u, v) \in E$ and a non-negative supply variable $f^s_v \in \mathbb{R}_{\geq 0}$ and demand variable $f^t_v \in \mathbb{R}_{\geq 0}$ for each vertex $v \in V$. We then formulate the constraints of the flow problem as follows:

\begin{align}
    \forall v \in V,\ f^t_v &= \sum_{x \in T} x_v
    \label{eq:connect-target}\\
    \forall v \in V,\ f^s_v &\leq |V| \cdot d_v \label{eq:connect-source}\\
    \forall v \in V,\ f^s_v + \sum_{u:(u,v) \in E} f(u,v) &= 
    f^t_v + \sum_{u:(v:u) \in E} f(u,v)\label{eq:flow-constraint}\\
    \forall (u, v) \in E,\ f(u,v) &\leq |V| \cdot (1 - \sum_{x \in B} x_u).
    \label{eq:block-flow}
\end{align}%
\Cref{eq:connect-target} guarantees that the demand $f^t_v$ is $1$ for sink vertices and $0$ for other vertices. 
Similarly, \Cref{eq:connect-source} guarantees that only the source vertex has a non-zero supply. We bound its supply by $|V|$ because each source vertex can reach at most $|V|$ sink vertices.
\Cref{eq:flow-constraint} is the flow conservation constraint, which ensures that the sum of the supply and the flow entering a vertex equals the sum of the demand and the flow leaving a vertex. \Cref{eq:block-flow} guarantees that no flow leaves a blocked vertex. 
Note that we do not need an objective in the flow problem formulation as our objective is defined in \Cref{eqn:obj}.

\subsubsection{Well-Form Constraints}

For the home-location scenario, the layouts must be well-formed for DPP to run. That is, we need to constrain that any two tiles of blue (endpoints) or orange (home locations) are connected through a path with only white tiles (empty spaces). 
To do so, we add the endpoints and home locations to the set of blocking tile types $B$, i.e., $B = \{s, h, e\}$. The other constraints do not change. Note that the reachability constraints are also more restrictive than the well-formed property, which allows, for example, an empty space to be surrounded by four endpoints. However, our setup rule out this case for simplicity.%

\subsection{Number of Shelves Constraints}

Finally, we constrain the number of shelves in the layout to be a given number $N_s$, which, in our experiment, is the number of shelves in the human-designed layouts of the same size. This constraint allows the optimized layouts to form a fair comparison with the human-designed layouts because we want both the optimized and human-designed warehouses to store the same number of shelves. In particular, we add the following constraint:

\begin{equation}
    \sum_{v \in V} s_v = N_s.
\end{equation}

\begin{table*}[!t]
    \centering
    \resizebox{\linewidth}{!}{
        \begin{tabular}{c|c|c|c|c}
        Setup & Archive Dim & Downsampled Archive Dim & Connected Shelf Components Range & Mean Task Length Range\\
        \hline
        1 & $[15, 100]$ & $[15, 25]$ & $[5, 20]$ & $[9, 14]$ \\
        2 & $[30, 100]$ & $[15, 25]$ & $[10, 40]$ & $[12, 18]$\\
        3 & $[100, 100]$ & $[20, 20]$ & $[140, 240]$ & $[27, 33]$\\
        4 & $[15, 100]$ & $[15, 25]$ & $[5, 20]$ & $[6, 12]$\\
        \end{tabular}
    }
    \caption{Hyperparameter of archives for the QD algorithm setup. Dim is short for dimensions.}
    \label{tab:add-param}
\end{table*}

\begin{figure}[!t]
    \begin{subfigure}[t]{0.24\textwidth}
        \centering
        \includegraphics[width=1\textwidth]{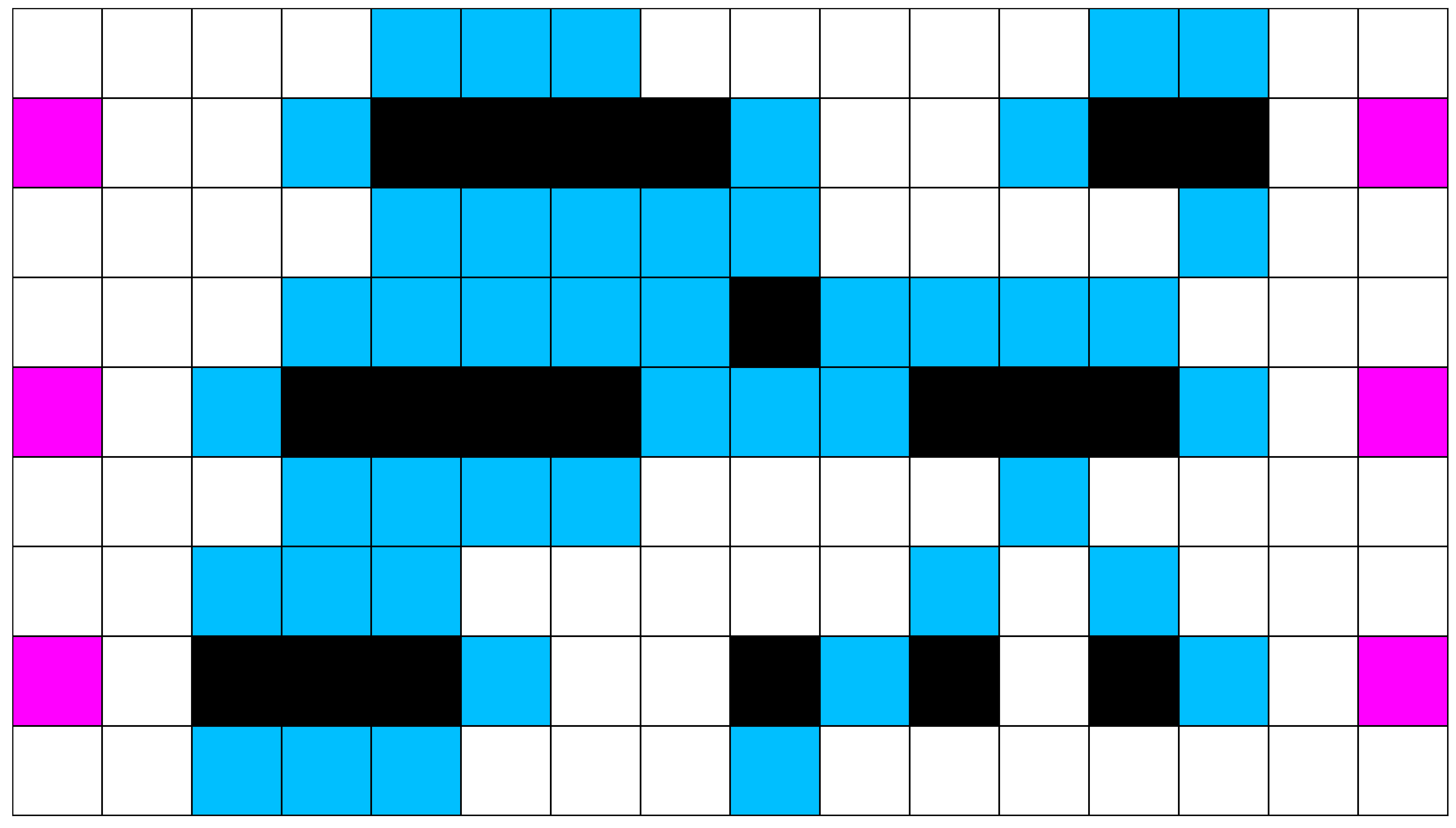}
        \caption{Setup 2 with DSAGE + RHCR}
        \label{fig:opt-small-w-RHCR-DSAGE}
    \end{subfigure}%
    \hfill
    \begin{subfigure}[t]{0.24\textwidth}
        \centering
        \includegraphics[width=1\textwidth]{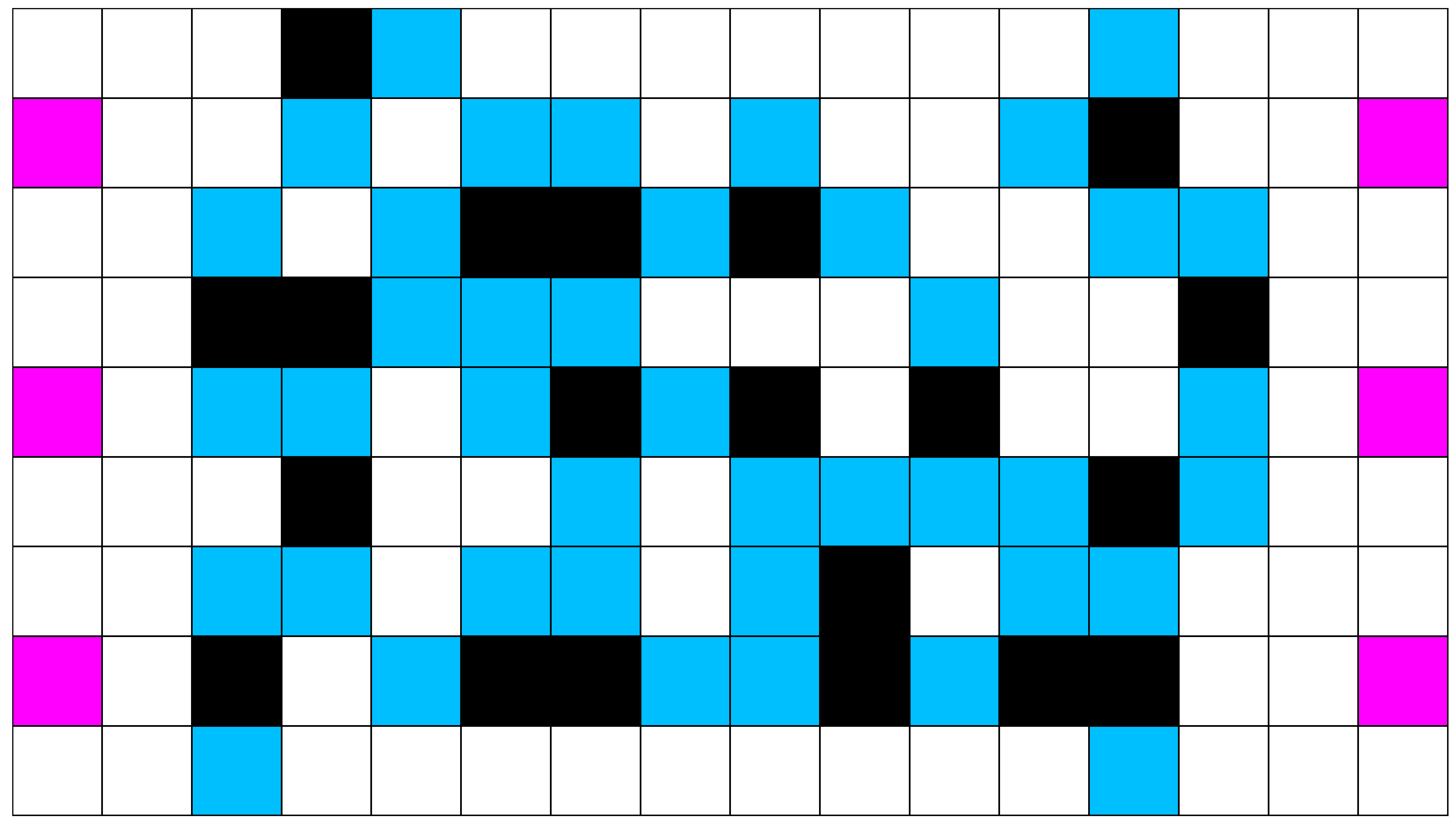}
        \caption{Setup 2 with MAP-Elites + RHCR}
        \label{fig:opt-small-w-RHCR-ME}
    \end{subfigure}\\
    \begin{subfigure}[t]{0.24\textwidth}
        \centering
        \includegraphics[width=1\textwidth]{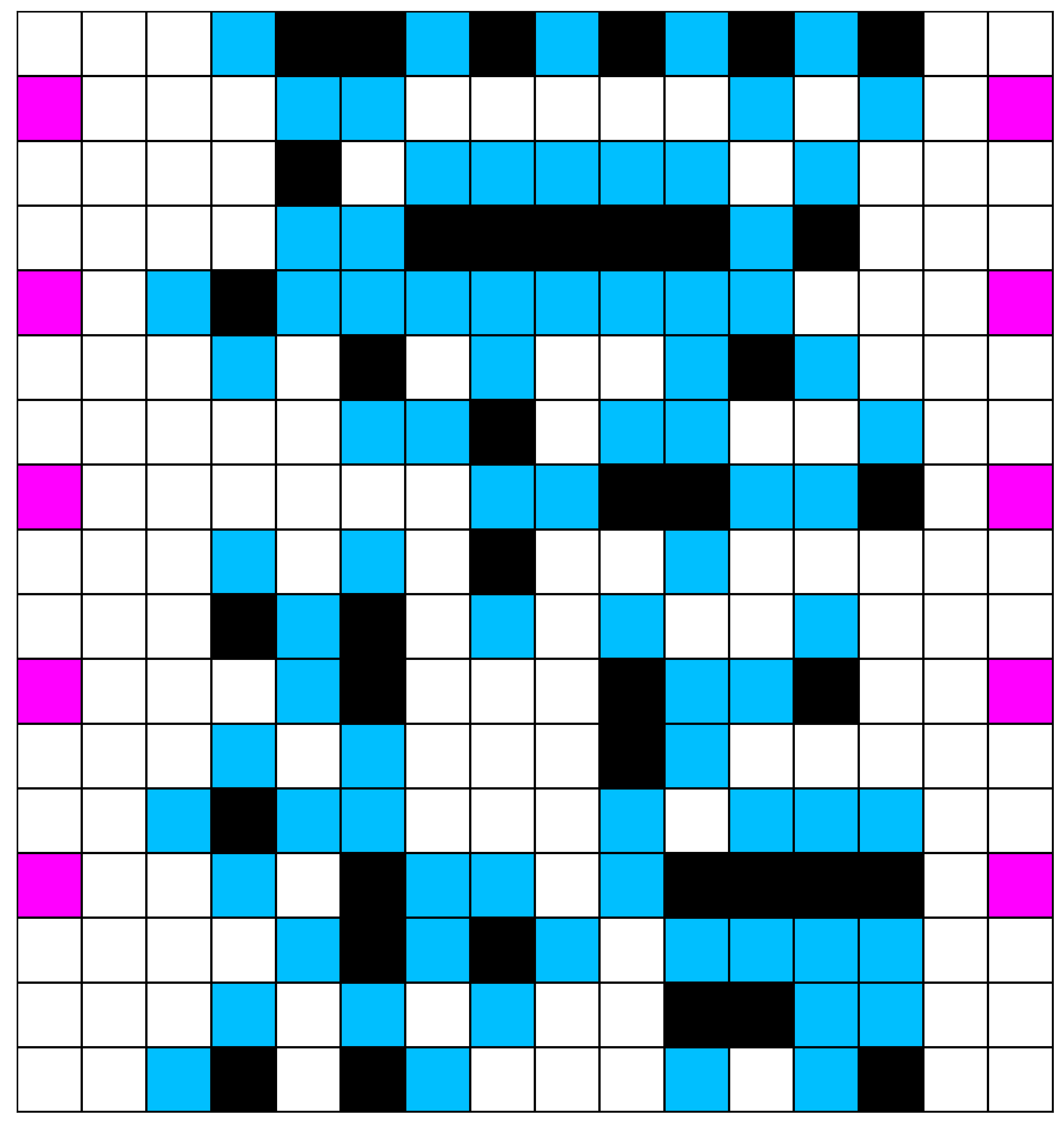}
        \caption{Setup 3 with DSAGE + RHCR}
        \label{fig:opt-medium-w-RHCR-DSAGE}
    \end{subfigure}%
    \hfill
    \begin{subfigure}[t]{0.24\textwidth}
        \centering
        \includegraphics[width=1\textwidth]{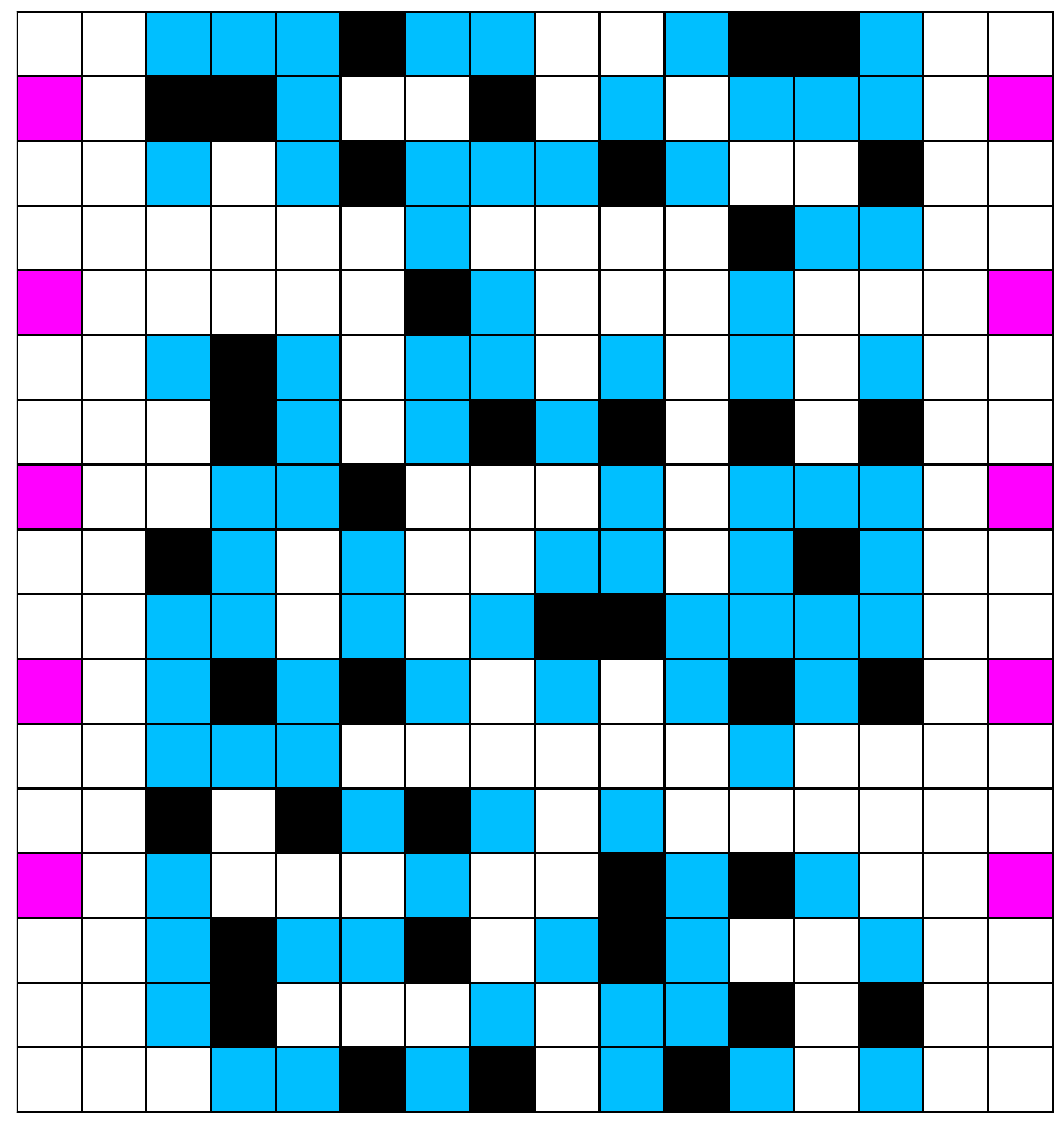}
        \caption{Setup 3 with MAP-Elites + RHCR}
        \label{fig:opt-medium-w-RHCR-ME}
    \end{subfigure}\\
    \hfill
    \begin{subfigure}[t]{0.24\textwidth}
        \centering
        \includegraphics[width=1\textwidth]{maps/global_opt/kiva_large_RHCR_w_200_agents_opt_dsage.png}
        \caption{Setup 4 with DSAGE + RHCR}
        \label{fig:opt-large-w-RHCR-DSAGE}
    \end{subfigure}%
    \hfill
    \begin{subfigure}[t]{0.24\textwidth}
        \centering
        \includegraphics[width=1\textwidth]{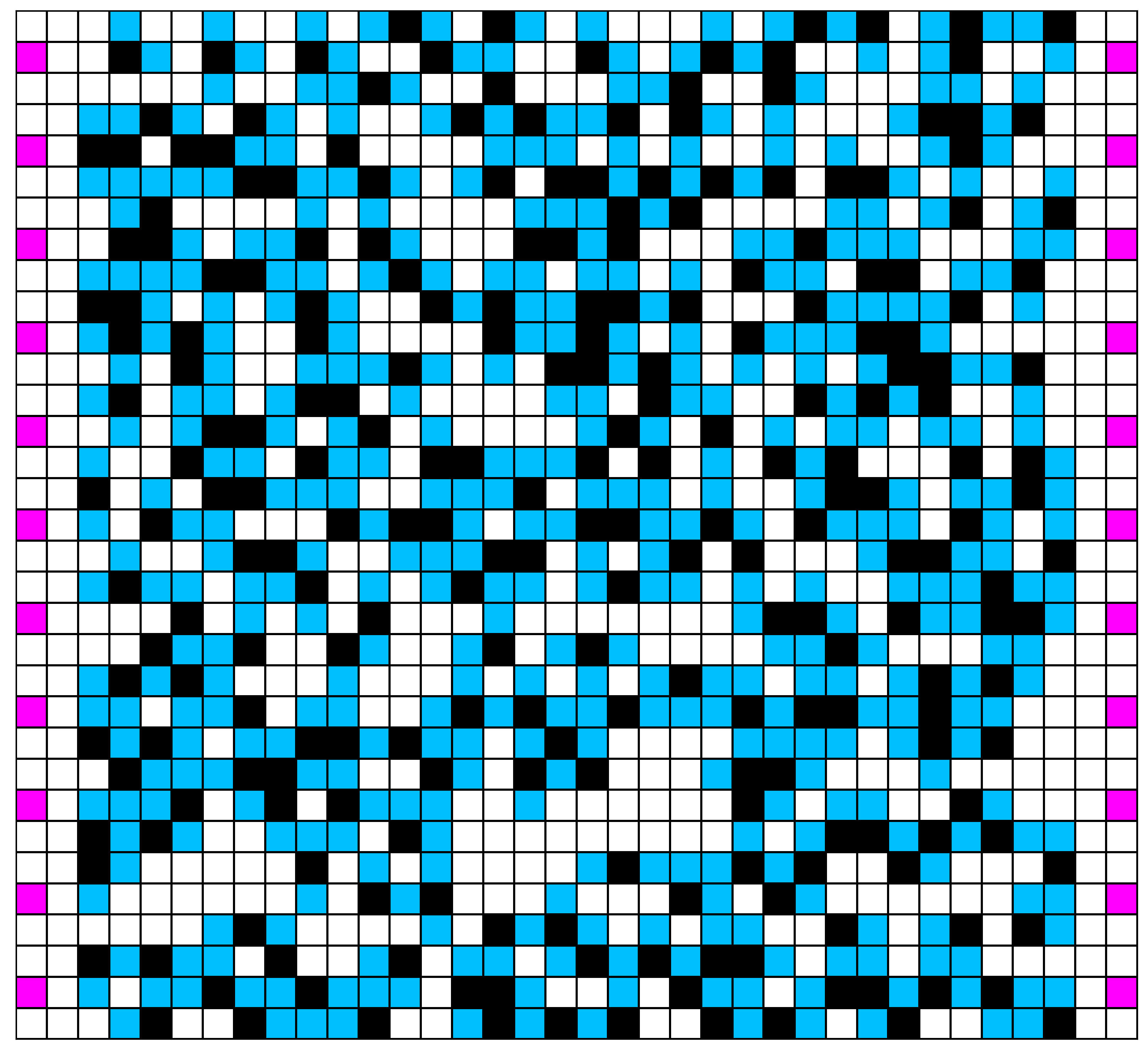}
        \caption{Setup 4 with MAP-Elites + RHCR}
        \label{fig:opt-large-w-RHCR-ME}
    \end{subfigure}
    \caption{Optimal optimized layouts in workstation scenario}
    \label{fig:optimal-w}
\end{figure}

\begin{figure}[!t]
    \begin{subfigure}[t]{0.23\textwidth}
        \centering
        \includegraphics[width=1\textwidth]{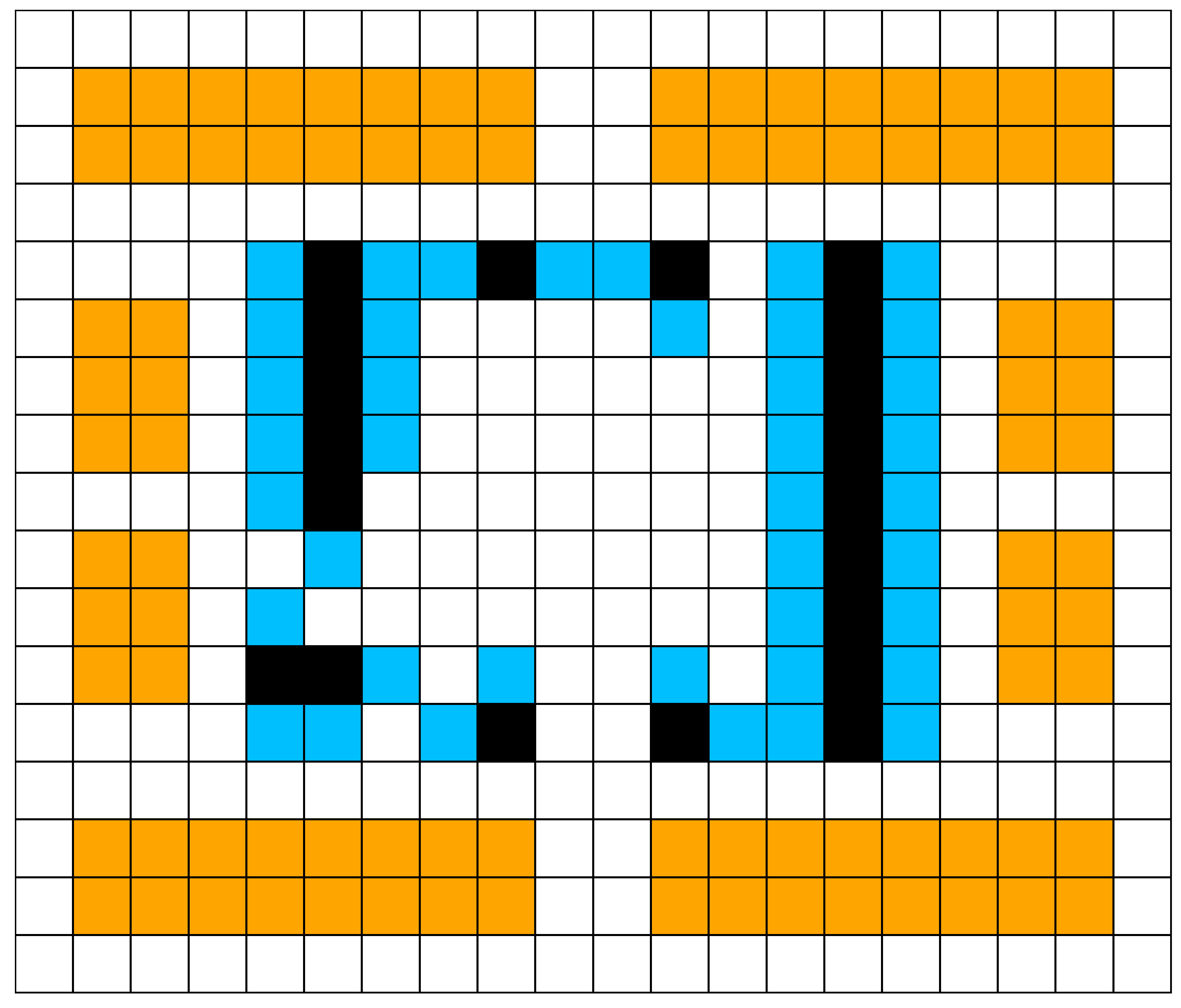}
        \caption{Setup 1 with DSAGE + RHCR and $88$ agents}
        \label{fig:opt-small-r-RHCR-DSAGE-88}
    \end{subfigure}%
    \hfill
    \begin{subfigure}[t]{0.23\textwidth}
        \centering
        \includegraphics[width=1\textwidth]{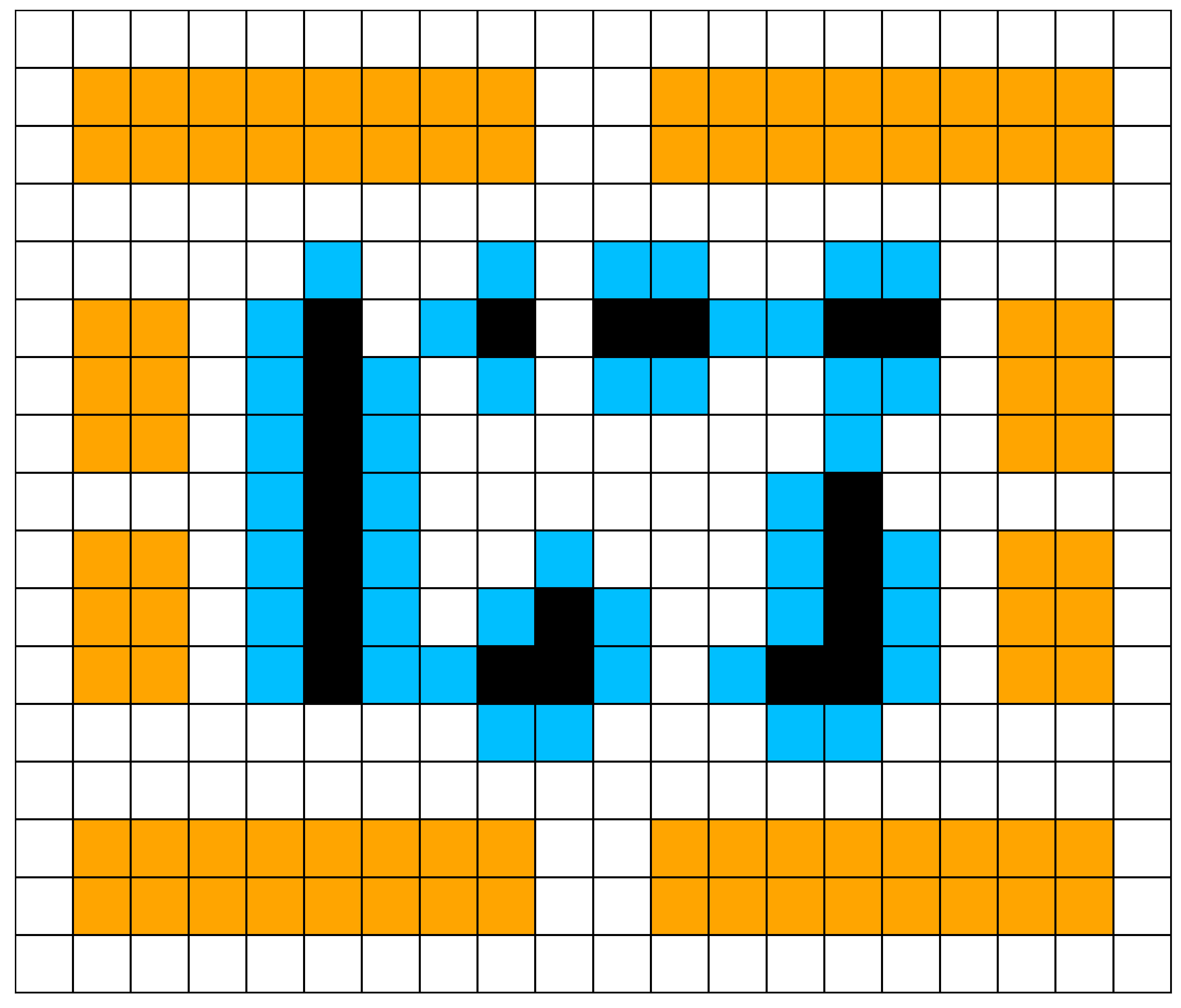}
        \caption{Setup 1 with DSAGE + RHCR and $60$ agents}
        \label{fig:opt-small-r-RHCR-DSAGE-60}
    \end{subfigure}\\
    \begin{subfigure}[t]{0.23\textwidth}
        \centering
        \includegraphics[width=1\textwidth]{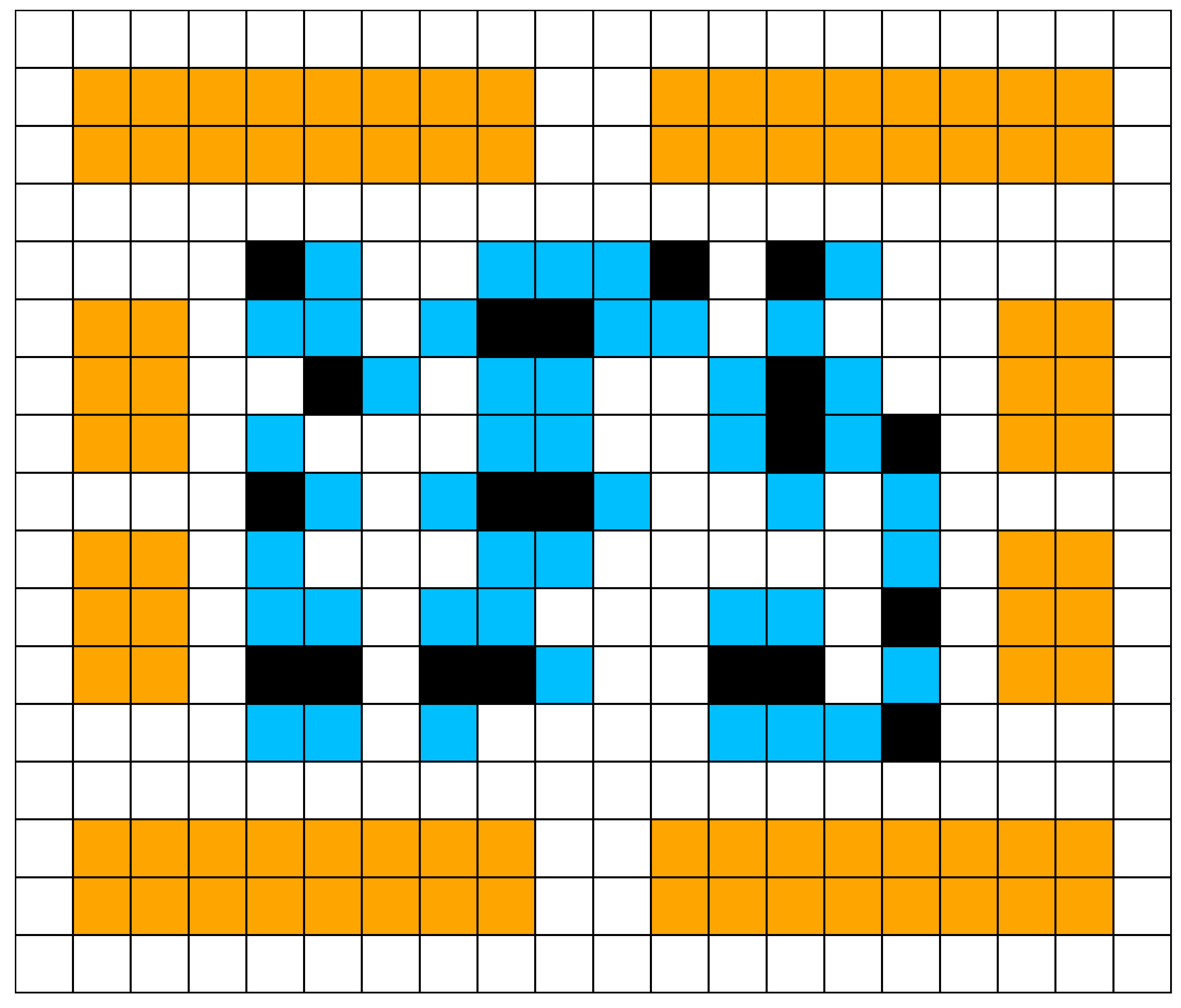}
        \caption{Setup 1 with DSAGE + RHCR and $40$ agents}
        \label{fig:opt-small-r-RHCR-DSAGE-40}
    \end{subfigure}
    \hfill
    \begin{subfigure}[t]{0.23\textwidth}
        \centering
        \includegraphics[width=1\textwidth]{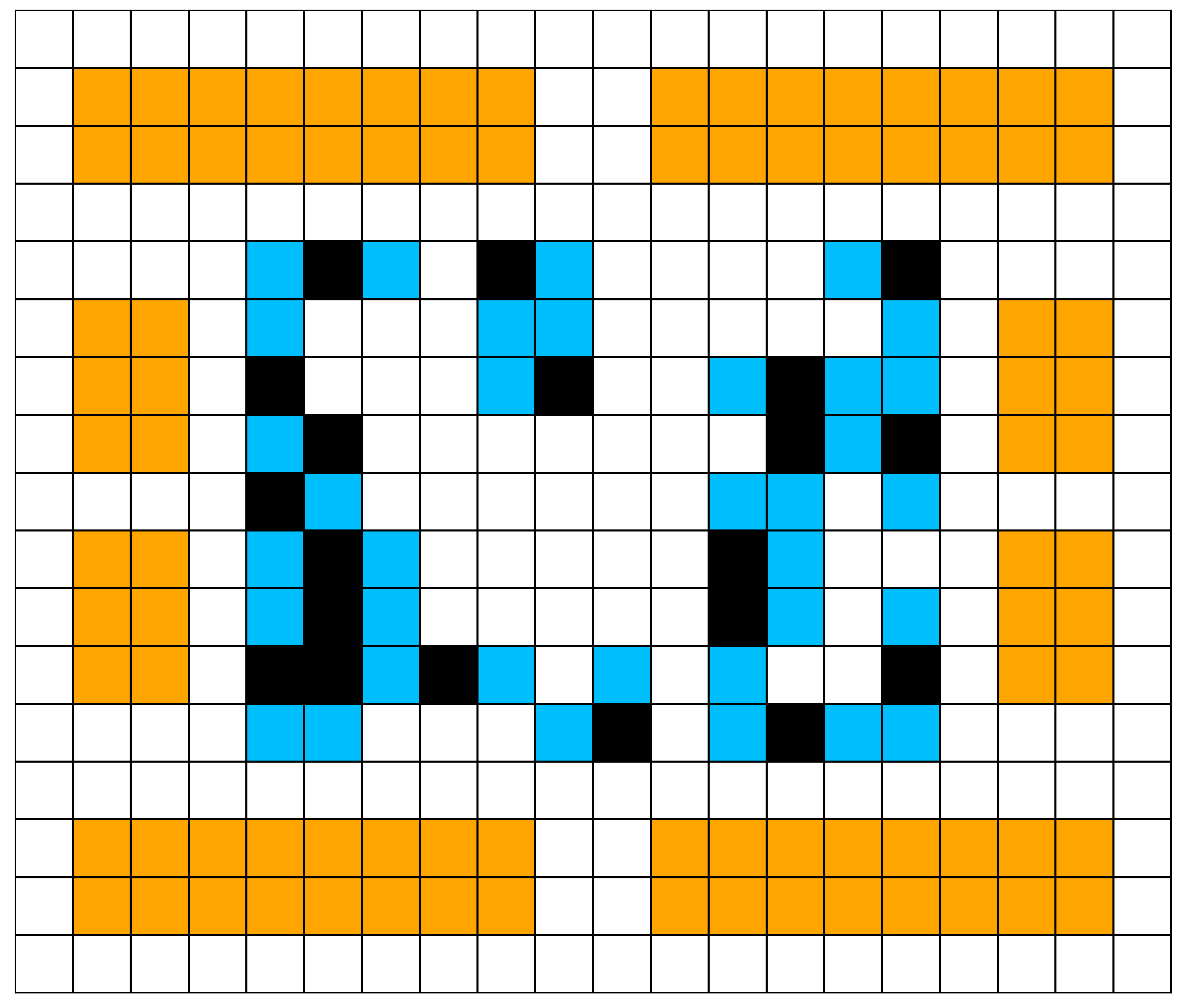}
        \caption{Setup 1 with MAP-Elites + RHCR}
        \label{fig:opt-small-r-RHCR-ME}
    \end{subfigure}\\
    \begin{subfigure}[t]{0.23\textwidth}
        \centering
        \includegraphics[width=1\textwidth]{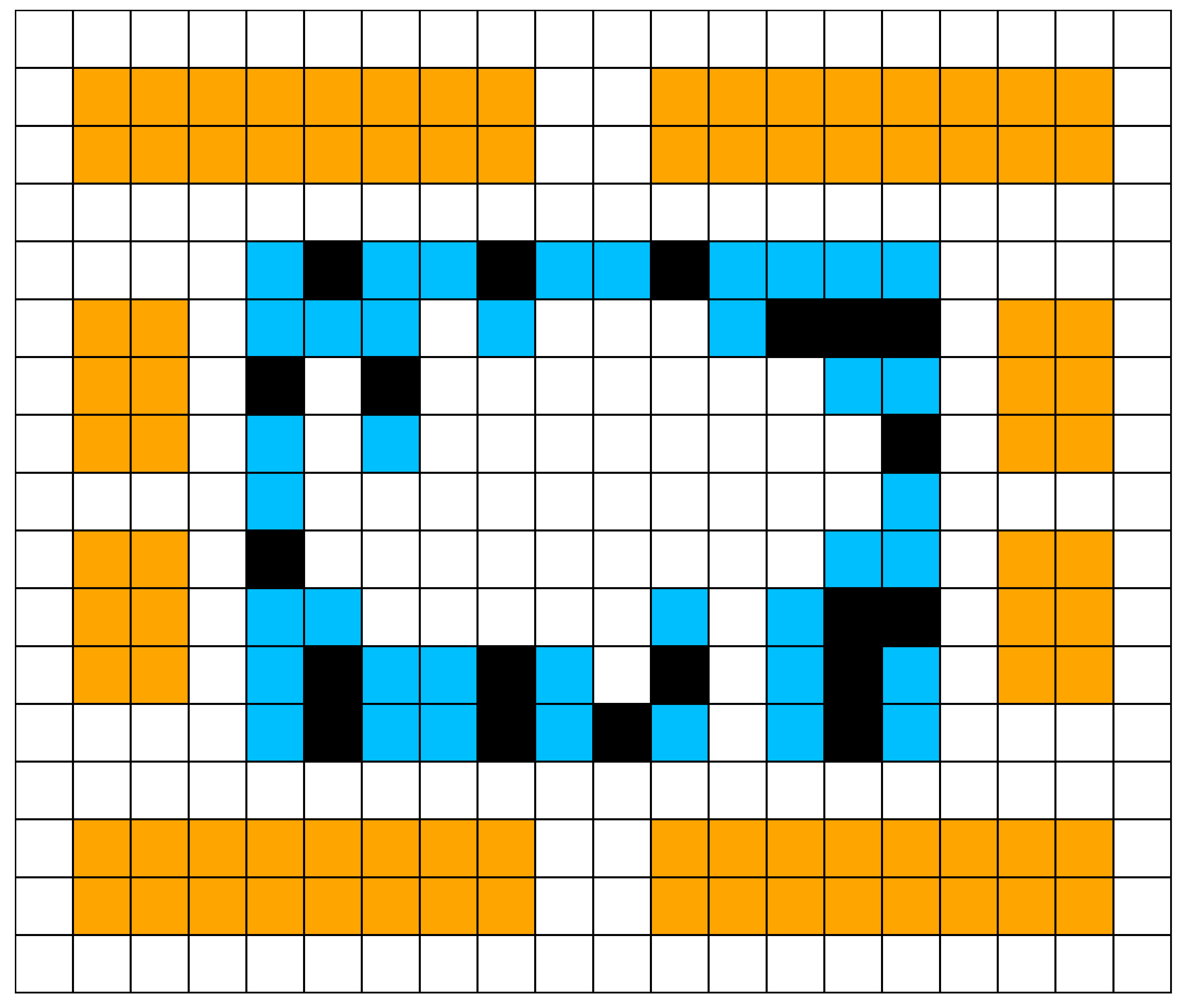}
        \caption{Setup 1 with DSAGE + DPP}
        \label{fig:opt-small-r-DPP-DSAGE}
    \end{subfigure}%
    \hfill
    \begin{subfigure}[t]{0.23\textwidth}
        \centering
        \includegraphics[width=1\textwidth]{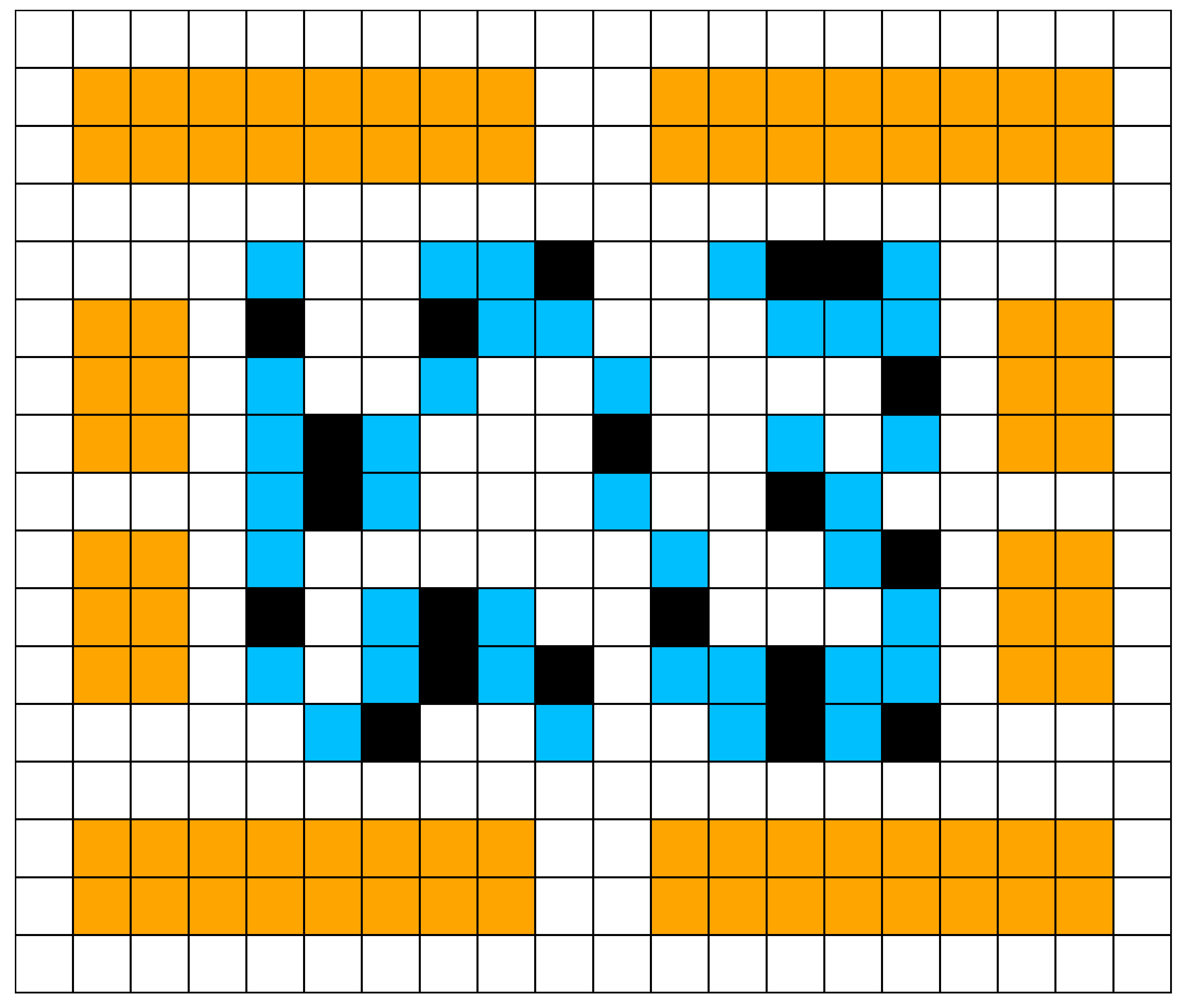}
        \caption{Setup 1 with MAP-Elites + DPP}
        \label{fig:opt-small-r-DPP-ME}
    \end{subfigure}%
    \caption{Optimal optimized layouts in home-location scenario}
    \label{fig:optimal-r}
\end{figure}

\subsection{Implementation}

We implement the MILP with IBM's CPLEX library~\cite{ibm_cplex} in Python.

\section{QD Optimization Setup} \label{appen:QD}

\begin{figure}[!t]
    \begin{subfigure}[t]{0.24\textwidth}
        \centering
        \includegraphics[width=1\textwidth]{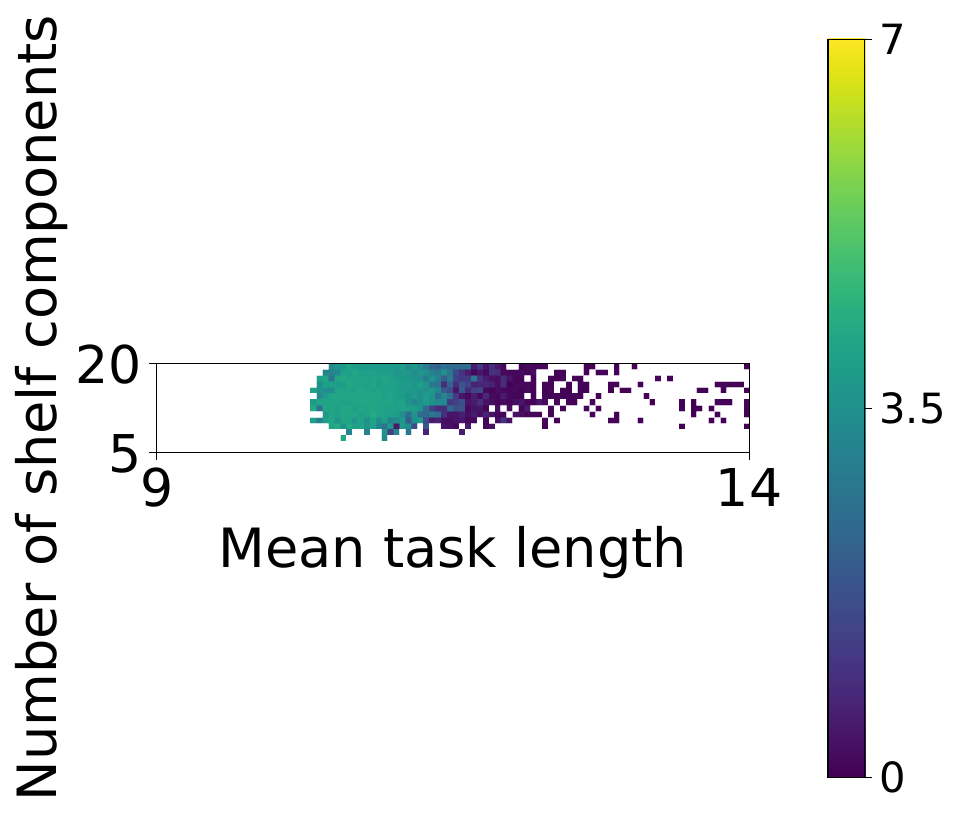}
        \caption{Setup 2 with DSAGE + RHCR}
        \label{fig:archive-small-w-RHCR-DSAGE}
    \end{subfigure}%
    \hfill
    \begin{subfigure}[t]{0.24\textwidth}
        \centering
        \includegraphics[width=1\textwidth]{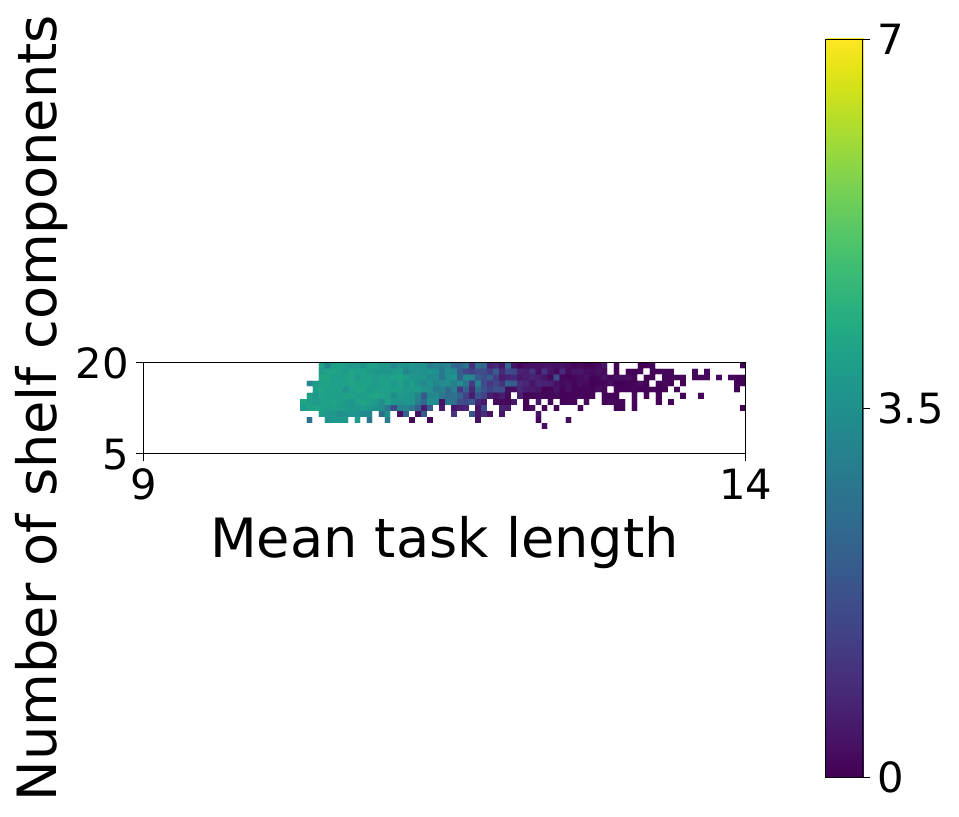}
        \caption{Setup 2 with MAP-Elites + RHCR}
        \label{fig:archive-small-w-RHCR-ME}
    \end{subfigure}\\
    \begin{subfigure}[t]{0.24\textwidth}
        \centering
        \includegraphics[width=1\textwidth]{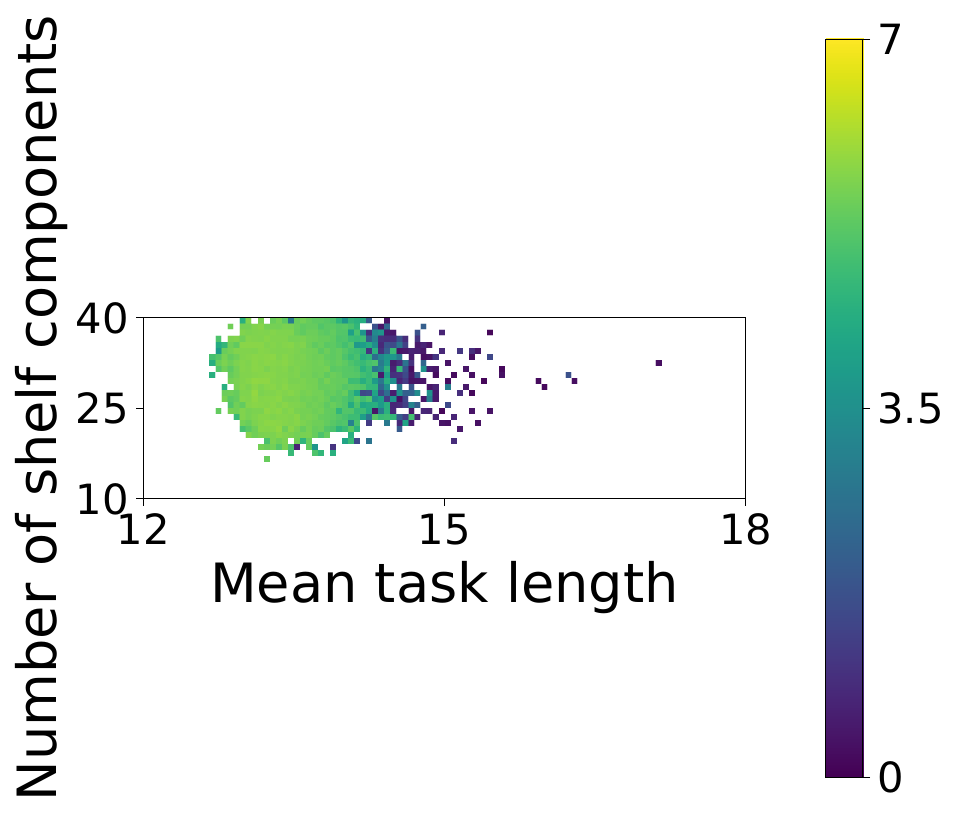}
        \caption{Setup 3 with DSAGE + RHCR}
        \label{fig:archive-medium-w-RHCR-DSAGE}
    \end{subfigure}%
    \hfill
    \begin{subfigure}[t]{0.24\textwidth}
        \centering
        \includegraphics[width=1\textwidth]{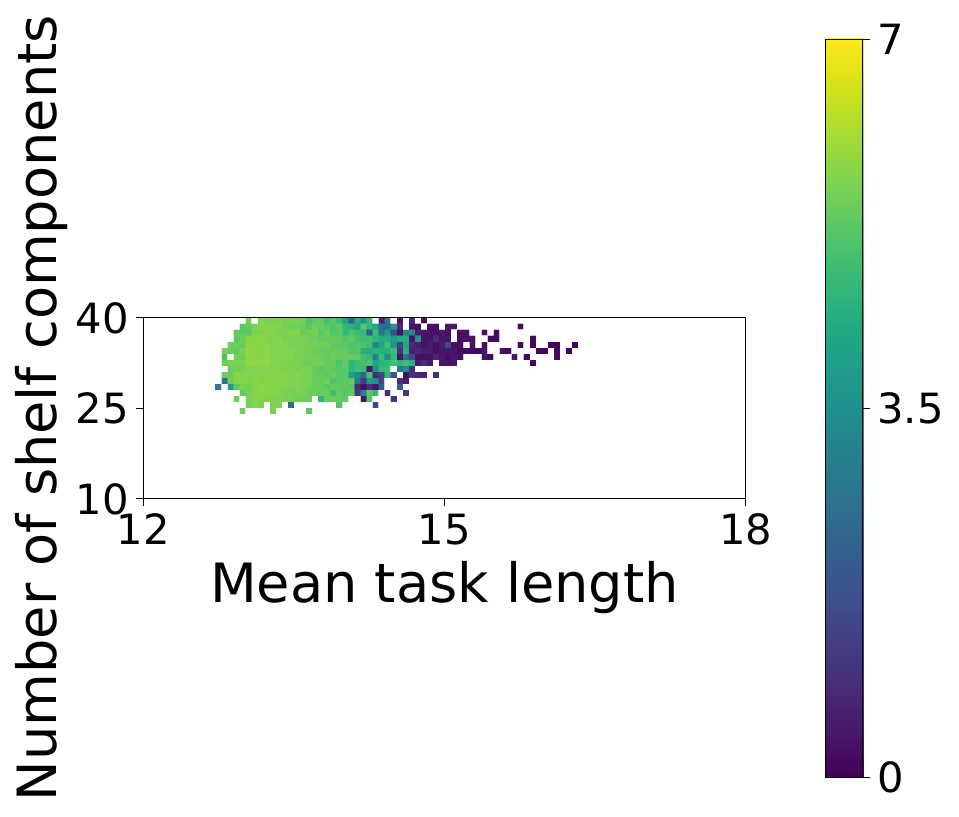}
        \caption{Setup 3 with MAP-Elites + RHCR}
        \label{fig:archive-medium-w-RHCR-ME}
    \end{subfigure}\\
    \begin{subfigure}[t]{0.24\textwidth}
        \centering
        \includegraphics[width=1\textwidth]{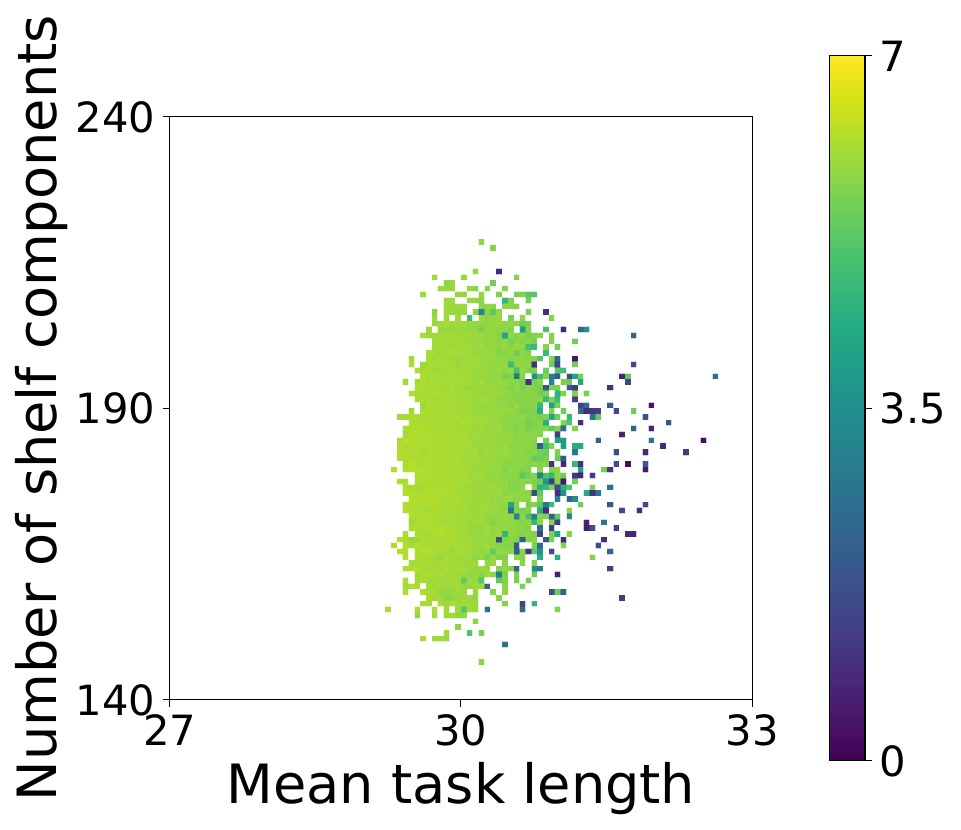}
        \caption{Setup 4 with DSAGE + RHCR}
        \label{fig:archive-large-w-RHCR-DSAGE}
    \end{subfigure}%
    \hfill
    \begin{subfigure}[t]{0.24\textwidth}
        \centering
        \includegraphics[width=1\textwidth]{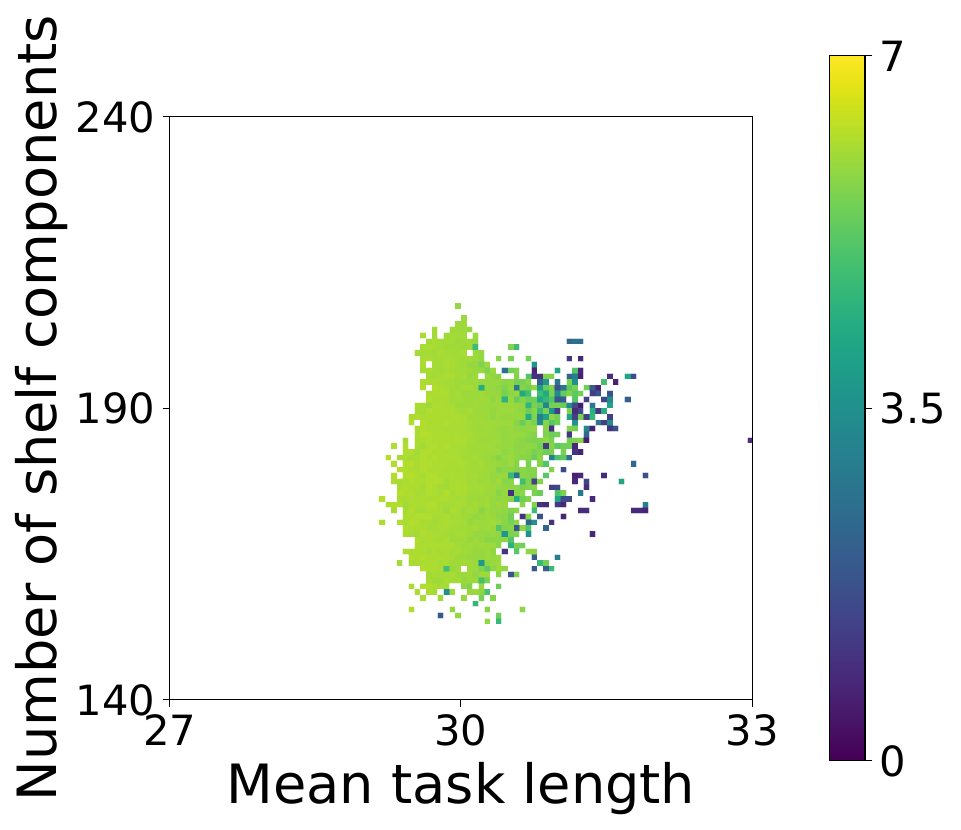}
        \caption{Setup 4 with MAP-Elites + RHCR}
        \label{fig:archive-large-w-RHCR-ME}
    \end{subfigure}
    \caption{Archives of workstation scenario}
    \label{fig:archive-optimal-w}
\end{figure}

\begin{figure}[!t]
    \begin{subfigure}[t]{0.24\textwidth}
        \centering
        \includegraphics[width=1\textwidth]{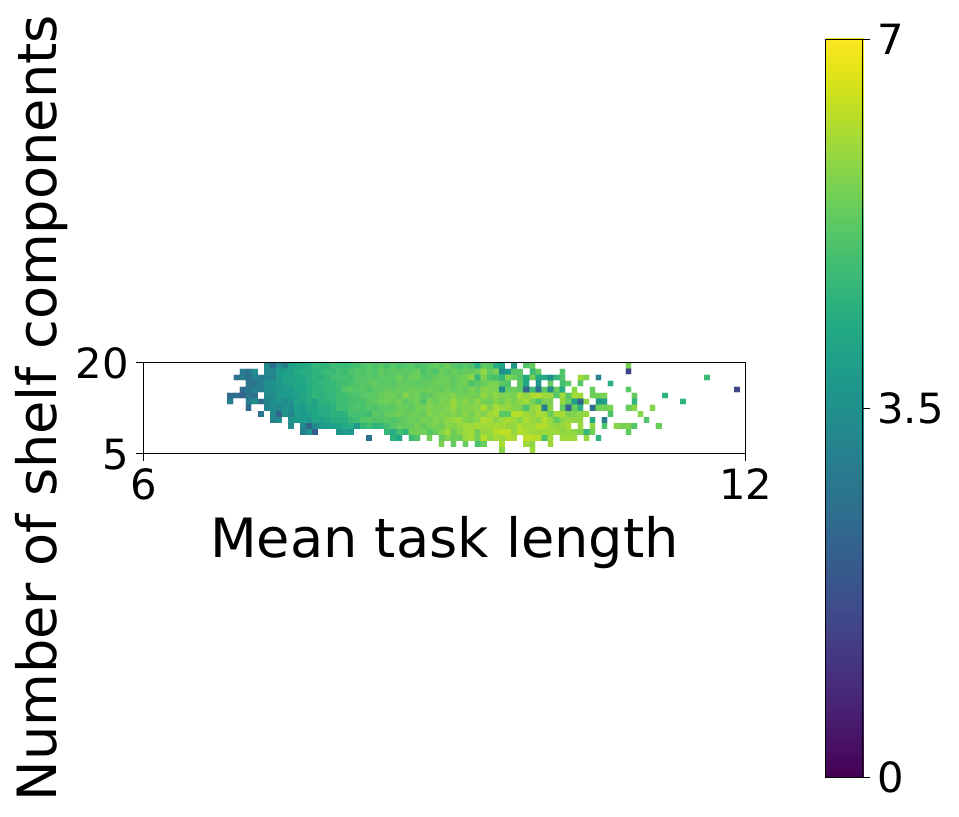}
        \caption{Setup 1 with DSAGE + RHCR and $88$ agents}
        \label{fig:archive-small-r-RHCR-DSAGE-88}
    \end{subfigure}%
    \hfill
    \begin{subfigure}[t]{0.24\textwidth}
        \centering
        \includegraphics[width=1\textwidth]{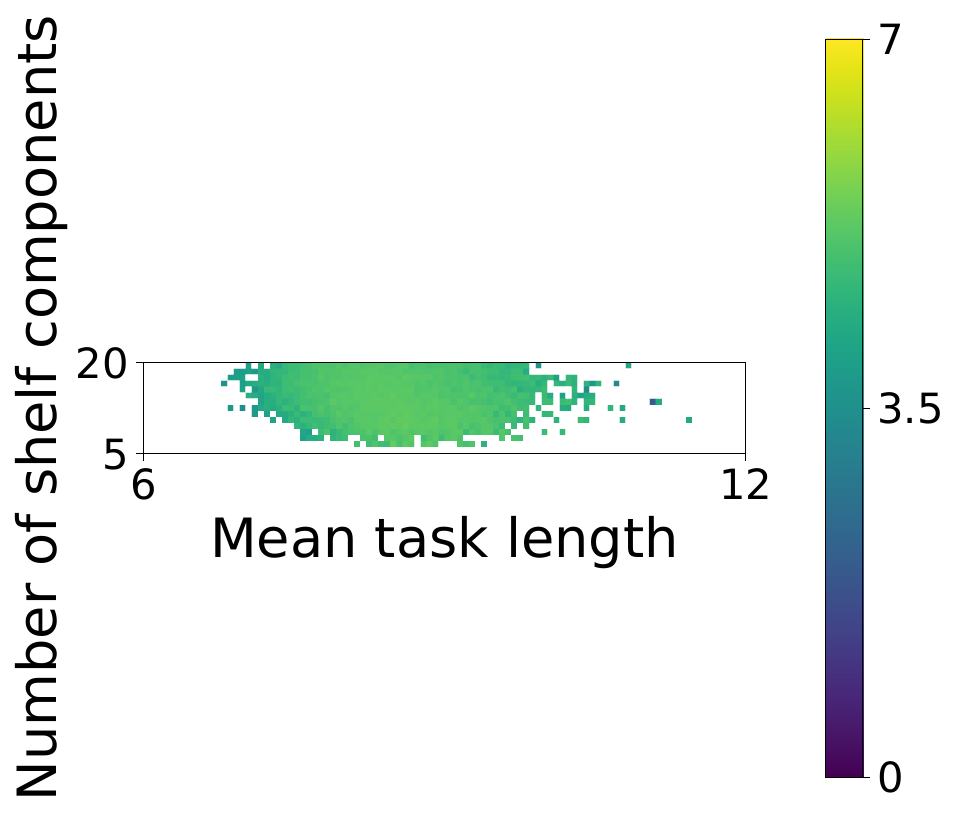}
        \caption{Setup 1 with DSAGE + RHCR and $60$ agents}
        \label{fig:archive-small-r-RHCR-DSAGE-60}
    \end{subfigure}\\
    \begin{subfigure}[t]{0.24\textwidth}
        \centering
        \includegraphics[width=1\textwidth]{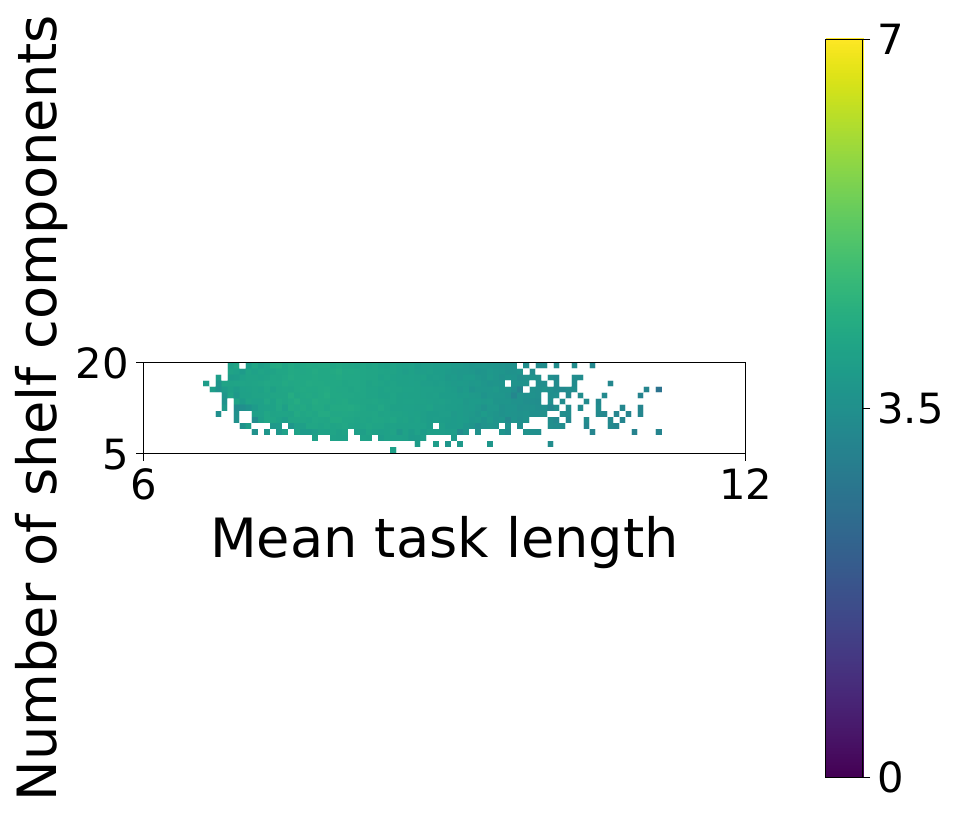}
        \caption{Setup 1 with DSAGE + RHCR and $40$ agents}
        \label{fig:archive-small-r-RHCR-DSAGE-40}
    \end{subfigure}%
    \hfill
    \begin{subfigure}[t]{0.24\textwidth}
        \centering
        \includegraphics[width=1\textwidth]{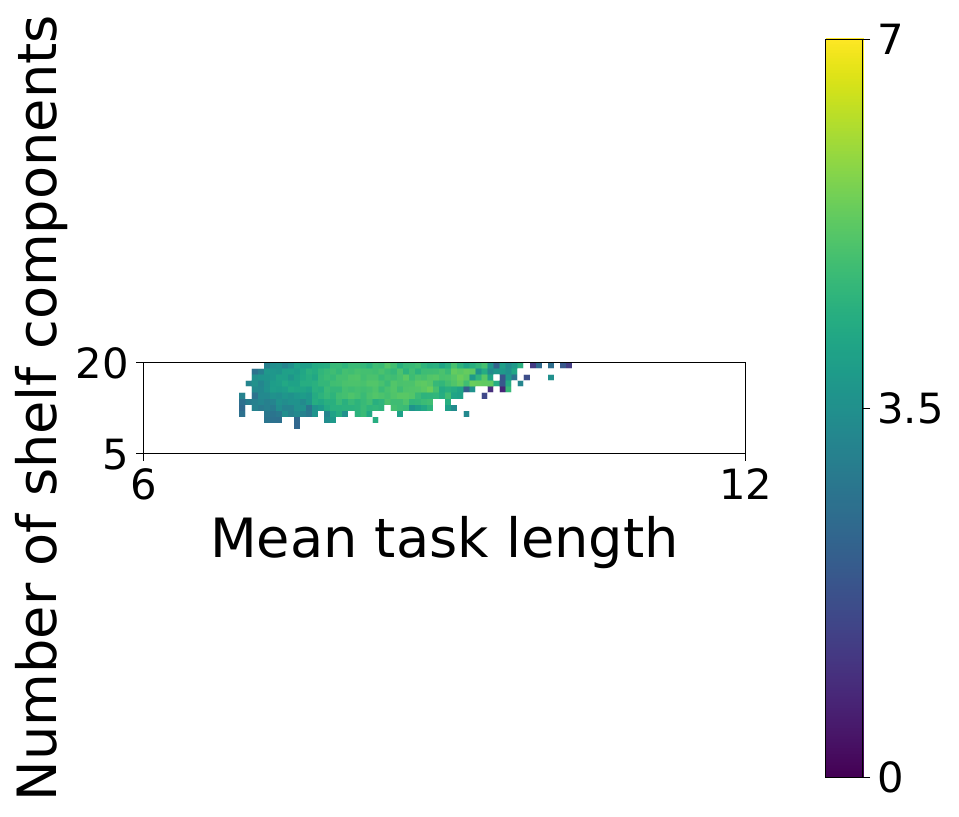}
        \caption{Setup 1 with MAP-Elites + RHCR}
        \label{fig:archive-small-r-RHCR-ME}
    \end{subfigure}\\
    \begin{subfigure}[t]{0.24\textwidth}
        \centering
        \includegraphics[width=1\textwidth]{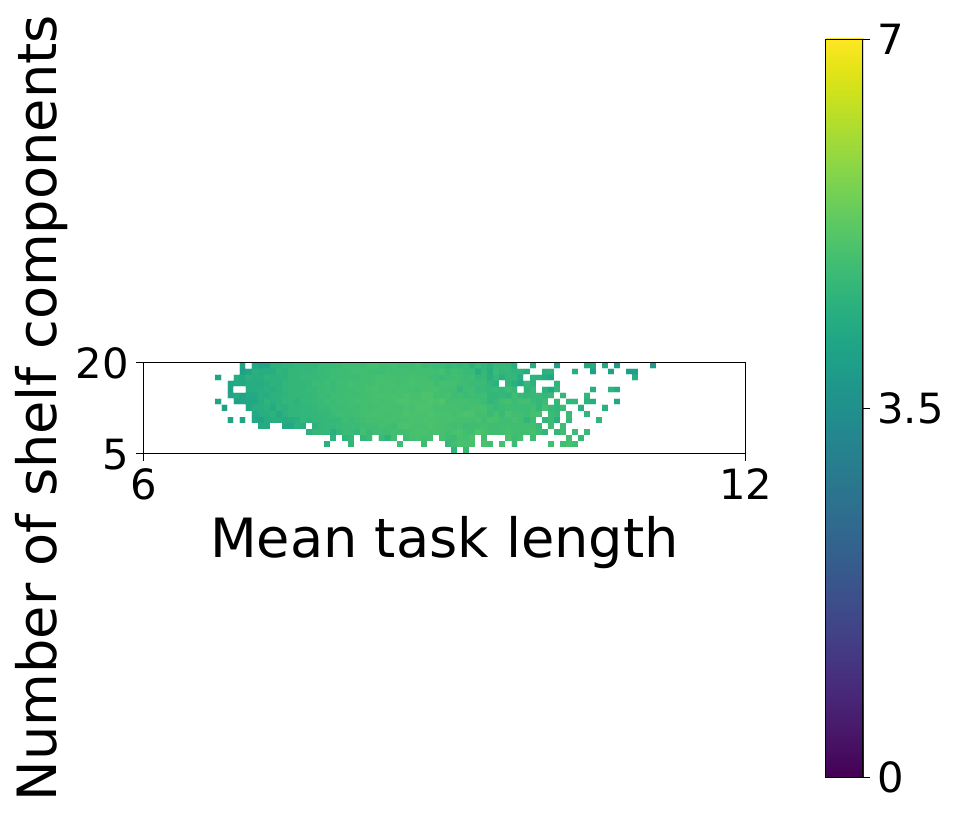}
        \caption{Setup 1 with DSAGE + DPP}
        \label{fig:archive-small-r-DPP-DSAGE}
    \end{subfigure}%
    \hfill
    \begin{subfigure}[t]{0.24\textwidth}
        \centering
        \includegraphics[width=1\textwidth]{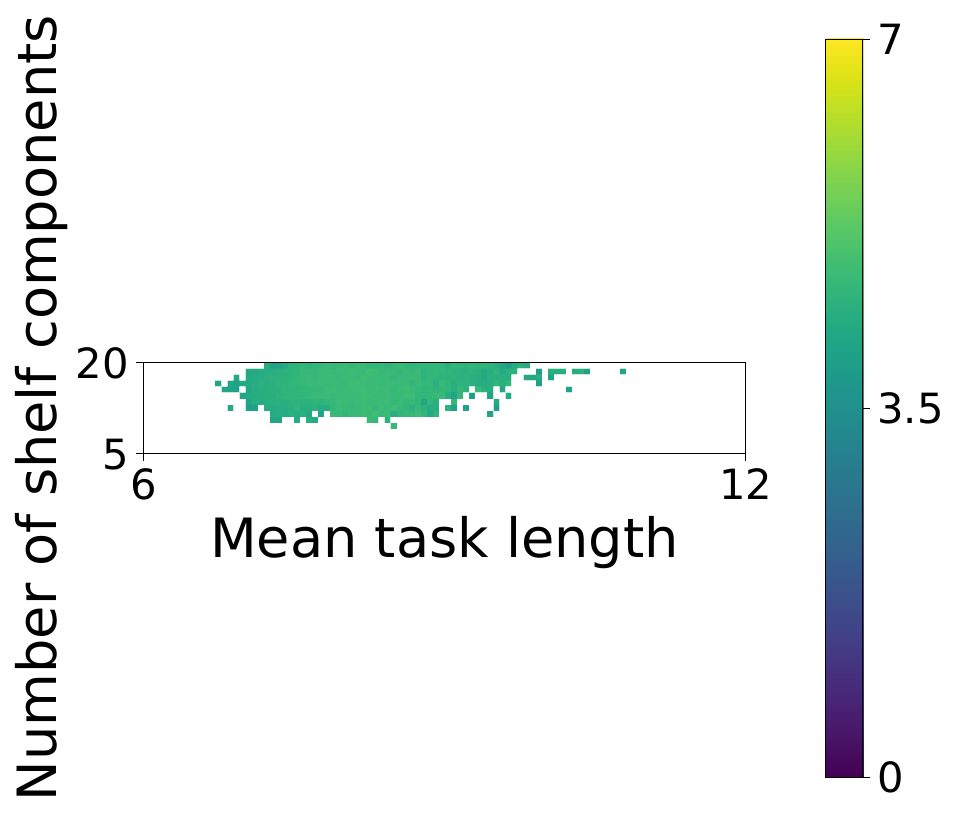}
        \caption{Setup 1 with MAP-Elites + DPP}
        \label{fig:archive-small-r-DPP-ME}
    \end{subfigure}%
    \caption{Archives in home-location scenario}
    \label{fig:archive-optimal-r}
\end{figure}

\subsection{Surrogate Model} \label{appen:subsec:model}

\paragraph{Model Architecture} \Cref{fig::surrogate-model-arch} shows the architecture of the surrogate model for setup 2 in \cref{sec:exp}. Following previous work~\cite{Bhatt2022DeepSA}, we use a three-stage deep convolutional neural network with three sub-networks $s_1$, $s_2$, and $s_3$ as the surrogate model of DSAGE.  Both sub-networks $s_1$ and $s_2$ use a convolution layer with a $3 \times 3$ kernel and a LeakyRELU activation function followed by $2$ residual layers~\cite{He2015DeepRL} and a convolution layer with $1 \times 1$ kernel. Sub-network $s_3$ starts with a convolution layer with a $4 \times 4$ kernel and stride $2$. Then we keep adding the same convolution layer with exponentially more channels until the height and width of the output are reduced to $4$. All convolutional layers in $s_3$ are followed by a batch normalization layer and a LeakyRELU activation function. Then, the output of the convolutions is flattened and passed through $2$ fully connected layers to predict the throughput and measures.

\paragraph{Layout Encoding} Following previous work~\cite{volz2018evolving}, we encode the layouts by mapping each tile to a distinct integer and then transforming each integer to a one-hot encoded vector. Therefore, each layout is represented by a $3$D tensor where each tile is a $1$D one-hot vector. 

\paragraph{Model Training} We train the sub-networks with different loss functions. %
Sub-network $s_1$ predicts the repaired layout, where the prediction for each tile can be viewed as a classification problem. We thus use the mean cross-entropy loss over all tiles as the loss function. Sub-network $s_2$ predicts the tile-usage map, which can be viewed as a distribution. We thus use the KL divergence loss as the loss function. Sub-network $s_3$ predicts the throughput and measures, which can be viewed as a regression problem. We thus use the mean squared error as the loss function. 
Since we use different loss functions with different ranges, we train each sub-network separately while freezing the other 2 sub-networks to prevent the gradients of any one loss function from dominating the others. We train each sub-network for $20$ epochs with a batch size of $64$. We use Adam optimizer~\cite{Kingma2014AdamAM} with  learning rate $lr = 0.001$, $\beta_1 = 0.9$, and $\beta_2 = 0.99$.

\subsection{Archive} \label{appen:subsec:archive}

\Cref{tab:add-param} shows the hyperparameters of the archives we used in \Cref{sec:exp}. The first column shows the archive dimension. The numbers in the bracket are the number of segments we discretize for the number of connected shelf components and the mean task length, respectively. The second column shows the dimension of the downsampled archive. The last two columns show the range of the number of connected shelf components and mean task length. The upper bound of the number of connected shelf components is always $N_s$ for each setup. We empirically decide the all the hyperparameters of the archive so that the measure space can cover most of the generated layouts.

\subsection{Computational Resource} \label{append:subsec:compute}

We run our experiments on five machines, namely (1) a local machine with AMD Ryzen $9$ $5950$X CPU, $32$ GB of RAM, and Nvidia Geforce RTX $3080$ GPU, (2) a local machine with AMD Ryzen Threadripper $3990$X, $64$ GB of RAM, and Nvidia Geforce RTX $3090$Ti GPU, 
(3) a local machine with AMD Ryzen Threadripper $3990$X, $64$ GB of RAM, and Nvidia Geforce RTX A$6000$ GPU, (4) a high-performing cluster with a V$100$ GPU and $200$ Xeon CPUs each with $4$ GB of RAM, and (5) a high-performing cluster with V$100$ GPU and heterogeneous CPUs. We measure the CPU runtime in \Cref{tab:numerical-result} with the first machine. We run our experiments on different machines because our experiments require massive parallelization, and all metrics except CPU runtime are not machine-sensitive.

The runtime of our QD algorithm for the home-location scenario ranges from $5$ hours (for DSAGE plus DPP) to $77$ hours (for MAP-Elites plus RHCR). For the workstation scenario, the runtime ranges from $12$ hours (for small layouts) to $24$ hours (for large layouts).

\subsection{Implementation}

The surrogate model is implemented using PyTorch~\cite{Paszke2019PyTorchAI} and the QD algorithms are implemented using Pyribs~\cite{pyribs}.

\section{Additional Results} \label{appen:archive}

\begin{figure}[!t]
    \centering
    \begin{subfigure}{0.44\textwidth}
        \centering
        \includegraphics[width=1\textwidth]{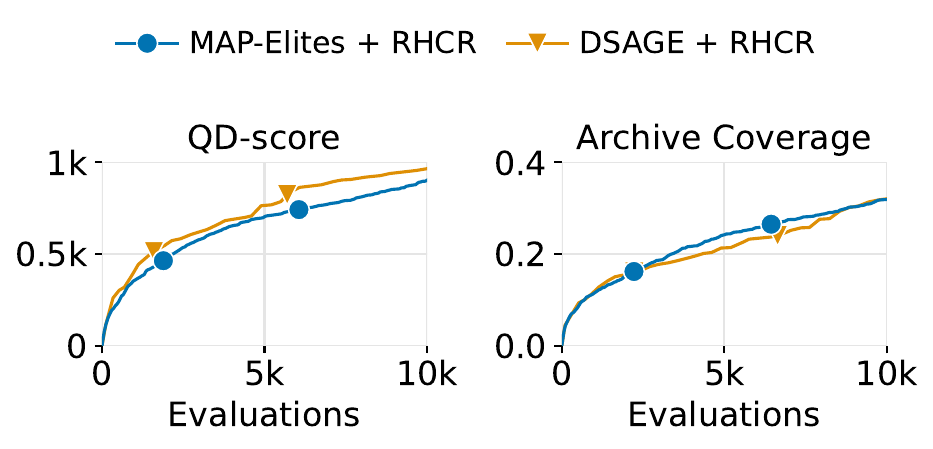}
        \caption{Setup 2}
        \label{fig:QD-score-small-w}
    \end{subfigure}\\
    \begin{subfigure}{0.44\textwidth}
        \centering
        \includegraphics[width=1\textwidth]{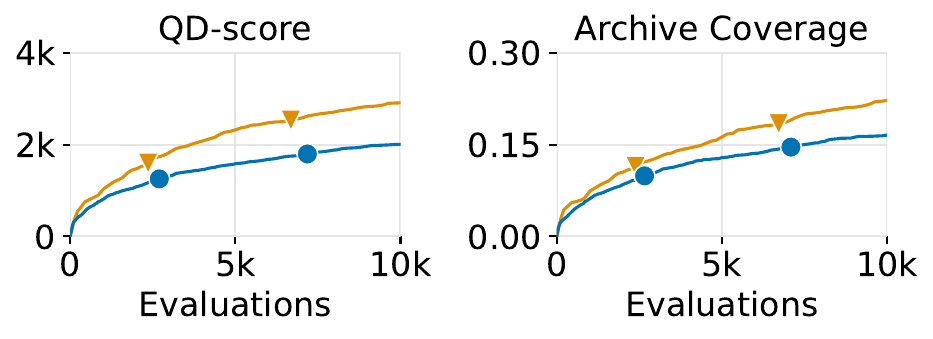}
        \caption{Setup 3}
        \label{fig:QD-score-mid-w}
    \end{subfigure}\\
    \begin{subfigure}{0.44\textwidth}
        \centering
        \includegraphics[width=1\textwidth]{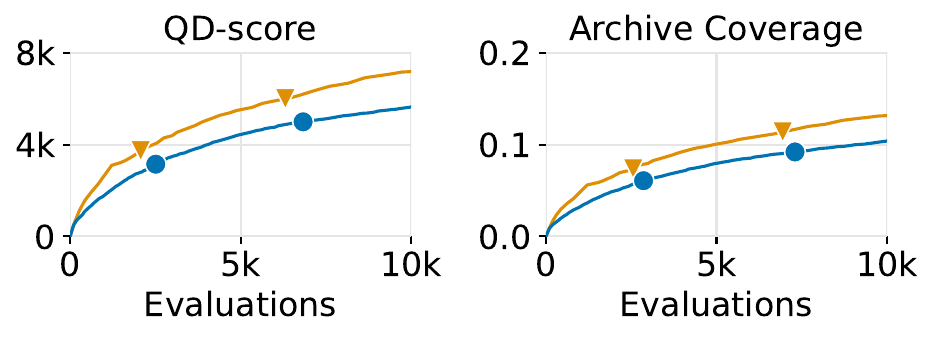}
        \caption{Setup 4}
        \label{fig:QD-score-large-w}
    \end{subfigure}
    \caption{QD-score and archive coverage of the archives for the workstation scenarios.}
    \label{fig:QD-score-w}
\end{figure}

\begin{figure}[!t]
    \centering
    \begin{subfigure}{0.45\textwidth}
        \centering
        \includegraphics[width=1\textwidth]{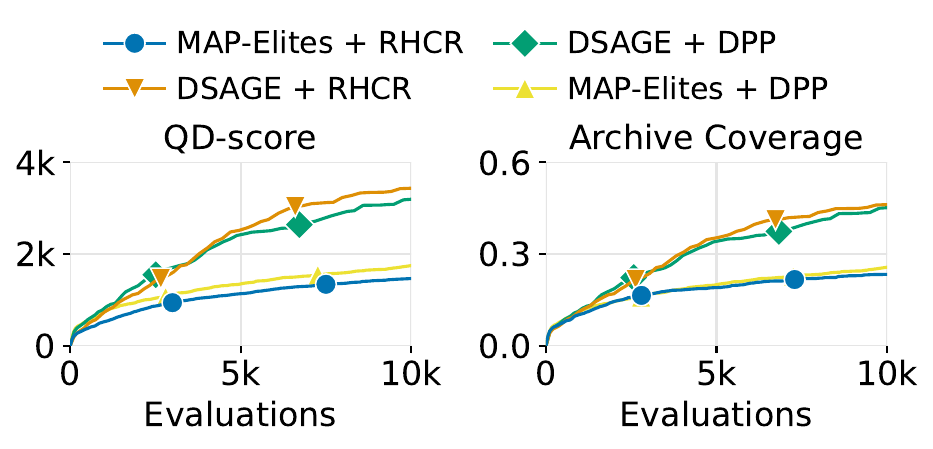}
        \caption{Setup 1}
        \label{fig:QD-score-small-r}
    \end{subfigure}\\
    \begin{subfigure}{0.48\textwidth}
        \centering
        \includegraphics[width=1\textwidth]{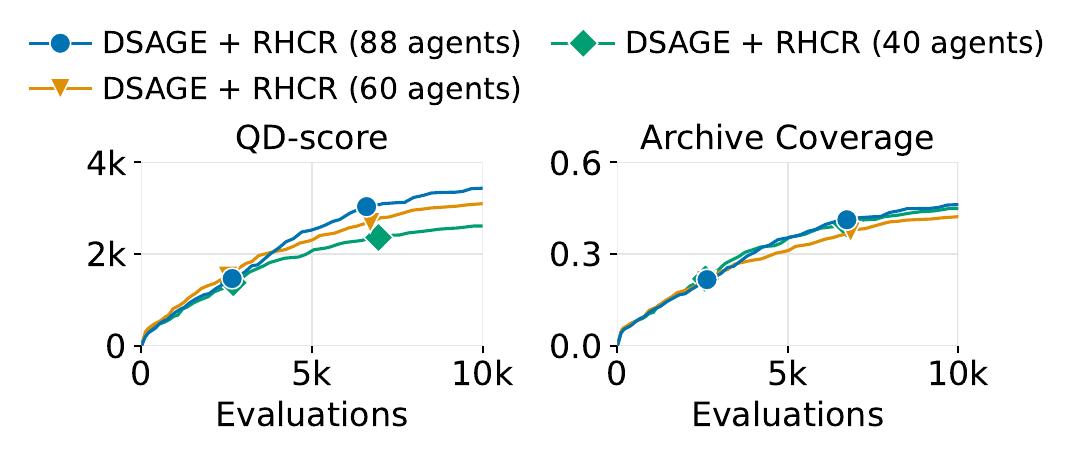}
        \caption{Setup 1 with different agents}
        \label{fig:QD-score-small-r-agents}
    \end{subfigure}
    \caption{QD-score and archive coverage of the archives for the home-location scenario.}
    \label{fig:QD-score-r}
\end{figure}

\subsection{Optimal Optimized Layouts}

\Cref{fig:optimal-w,fig:optimal-r} show the best optimized layouts in each experiment setup defined in \Cref{sec:exp}. We observe no clear regularized patterns in the workstation scenarios. However, in the home-location scenario (setup 1), we observe that the optimized layouts tend to distribute the shelves and endpoints along the borders of the storage area so that the agents can traverse among the endpoints through the empty spaces in the middle. Compared to MAP-Elites, DSAGE is more capable of finding this optimized pattern because DSAGE leverages a surrogate model to predict the output of the MILP solver and lifelong MAPF simulator on a given layout. The key insight behind the difference in performance is that MAP-Elites evaluates every candidate layout, whereas DSAGE evaluates only promising layouts obtained by exploiting the surrogate model.

\subsection{Archive and QD-score}

QD optimization attempts to maximize the throughput of the layout in each of the $M$ discretized cells in the archive. Formally, we maximize the QD-score~\cite{Justin2015confronting} $\sum_{j=1}^{M} f(\Vec{x}_j)$, where, in case no layout is found in a cell $j$, $f(\Vec{x}_j)$ is set to $0$. QD-score measures the overall quality and diversity of the layouts in the archive. In addition to QD-score, we also measure the archive coverage, which is the percentage of cells that contain an elite layout.

\Cref{fig:archive-optimal-w,fig:archive-optimal-r} show the archives of both MAP-Elites and DSAGE for all experiments in \Cref{sec:exp}. We qualitatively observe that DSAGE has better archive coverage than MAP-Elites, generating more layouts in the archive. \Cref{fig:QD-score-w,fig:QD-score-small-r} show the quantitative comparison of QD-scores and archive coverage for all experiments throughout $10,000$ evaluations on the lifelong MAPF simulator. Quantitatively, DSAGE achieves equal or higher QD-score and archive coverage than MAP-Elites in all setups. However, as shown in \Cref{fig:QD-score-small-w}, DSAGE does not show a clear advantage in setup 2, especially in terms of archive coverage. This is related to our discussion of the challenges of setup 2 in \Cref{subsec:result}. Due to the higher percentage of occupied traversable tiles by the agents ($P_a$), it is more challenging to improve the throughput by optimizing the layouts. We conjecture that the lifelong MAPF algorithm becomes the bottleneck of the throughput in setup 2. Therefore, optimizing layouts in setup 2 is more challenging than the other setups.

\Cref{fig:QD-score-small-r-agents} shows the QD-score and archive coverage of DSAGE for the home-location scenario with different numbers of agents. We observe that running DSAGE with different numbers of agents achieves similar archive coverage. However, running DSAGE with more agents results in better QD-scores because more agents can lead to better throughput for a given layout.

\end{document}